\documentclass[12pt]{iopart}
\usepackage[utf8]{inputenc}
\usepackage[]{graphicx}
\expandafter\let\csname equation*\endcsname=\relax
\expandafter\let\csname endequation*\endcsname=\relax
\usepackage{amssymb,amsmath}
\usepackage{stmaryrd}
\usepackage{bm}
\usepackage{subfigure}
\usepackage{cite}
\usepackage{algorithmic}
\usepackage{algorithm}
\usepackage{url}
\DeclareMathOperator{\argmin}{\text{argmin}}
\DeclareMathOperator{\argmax}{\text{argmax}}
\DeclareMathOperator{\dom}{dom}
\DeclareMathOperator{\prox}{prox}

\DeclareMathOperator{\ST}{\mathrm{SoftThresh}}

\DeclareMathOperator*{\s.t.}{s.t.}
\DeclareMathOperator*{\var}{Var}
\usepackage{mathtools}

\DeclarePairedDelimiter\floor{\lfloor}{\rfloor}

\begin{document}

\title{Constraint matrix factorization for space variant PSFs field restoration}

\author{F. Ngolè$^1$, J.-L. Starck$^1$, K. Okumura$^1$, J. Amiaux$^1$ and P. Hudelot$^2$}

\address{$^1$Laboratoire AIM, CEA/DSM-CNRS-Universite Paris Diderot, Irfu, Service d'Astrophysique, CEA Saclay, Orme des Merisiers, 91191 Gif-sur-Yvette, France\\
$^2$Institut d’Astrophysique de Paris, UMR 7095 CNRS \& UPMC, 98 bis Boulevard Arago, F-75014 Paris, France}

\ead{ngolefred@yahoo.fr}
\vspace{10pt}
\begin{indented}
\item[]November 2015
\end{indented}

\begin{abstract}

\underline{Context}: in large-scale spatial surveys, the Point Spread Function (PSF) varies across the instrument field of view (FOV). Local measurements of the PSFs are given by the isolated stars images. Yet, these estimates may not be directly usable for post-processings because of the observational noise and potentially the aliasing.
  
\underline{Aims}:  given a set of aliased and noisy stars images from a telescope, we want to estimate well-resolved and noise-free PSFs at the observed stars positions, in particular, exploiting the spatial correlation of the PSFs across the FOV.

\underline{Contributions}: we introduce RCA (Resolved Components Analysis) which is a noise-robust dimension reduction and super-resolution method based on matrix-factorization. We propose an original way of using the PSFs spatial correlation in the restoration process through sparsity. The introduced formalism can be applied to correlated data sets with respect to any euclidean parametric space.

\underline{Results}: we tested our method on simulated monochromatic PSFs of Euclid telescope (launch planned for 2020). The proposed method outperforms existing PSFs restoration and dimension reduction methods. We show that a coupled sparsity constraint on individual PSFs and their spatial distribution yields a significant improvement on both the restored PSFs shapes and the PSFs subspace identification, in presence of aliasing. 

\underline{Perspectives}: RCA can be naturally extended to account for the wavelength dependency of the PSFs.
\end{abstract}

%
\vspace{2pc}

\section{Introduction}
\label{sec:intro}  

In many applications such as high precision astronomical imaging or biomedical imaging, the optical system introduces a blurring of the images that needs to be taken into
account for scientific analyses, and the blurring function, also called Point Spread Function (PSF), is not always stationary on the observed field of view (FOV). A typical example is the case of the Euclid space mission \cite{Eucl1}, to be launched in 2020, where we need to measure with a very high accuracy the shapes of more than one billion of galaxies.  An extremely important step to derive such measurements is to get an estimate of the PSF at any spatial position of the observed images. This makes the PSF modeling a critical task.
In first approximation, the PSF can be modeled as a convolution kernel which is typically space and time-varying.
Several works in image processing \cite{escande} and specifically in astronomy \cite{port_1,den}, address the general problem of restoring images in presence of a space variant blur, assuming that the convolution kernel is locally known.

In astronomical imaging, unresolved objects such as stars, 
can provide PSF measurements  at different locations in the FOV. 
Nevertheless, these images can be aliased given the CCD sensors sizes which makes a super-resolution (SR) step necessary. This is the case for instance for the Euclid mission.

The SR is a widely studied topic in general image processing literature\cite{sina}. In astronomy,  softwares IMCOM \cite{im_com} and PSFEx\cite{bertin}, which propose an SR option,
are widely used. The IMCOM provides an oversampled output image from multiple under-sampled input images, assuming that the PSF is perfectly known. 
It does not deal with the PSF restoration itself. The PSFEx treats SR as an inverse problem, with a quadratic regularizer. In \cite{fng},  a sparsity based super-resolution method
was proposed, assuming that several low resolution (LR) measurements of the same PSF are available. In practice, we generally don't have such multiple measurements.

In this paper, we consider the case where the PSF is both space variant and under-sampled, 
and we want to get an accurate modeling at high resolution of the PSF field, assuming we have under-sampled measurements of different PSFs in the observed field. We assume that the PSFs vary slowly across the field. Intuitively, this implies a compressibility of the PSFs field, which leads us to the question of what would be a concise and easily understandable representation of a spatially indexed set of PSFs. 
\section{Notations}
We adopt the following notation conventions:
\begin{itemize}
\item bold low case letters are used for vectors;
\item bold capital case letters are used for matrices;
\item we treat vectors as column vectors unless explicitly mentioned otherwise. 
\end{itemize}
For a matrix $\mathbf{M}$, we note $m_{ij}$ the $j^{th}$ coefficient of the $i^{th}$ line, $\mathbf{m}_j^{(c)}$ or $\mathbf{M}[:,j]$ its $j^{th}$ column and $\mathbf{m}_i^{(l)}$ or $\mathbf{M}[i,:]$ its $i^{th}$ line, that we treat as a line vector. More generally for $j_1 \leq j_2$, we note $\mathbf{M[:,j1:j2]}$ the matrix obtained by extracting the columns of $\mathbf{M}$ indexed from $j_1$ to $j_2$; for $i_1 \leq i_2$, $\mathbf{M[i1:i2,:]}$ is defined analogously with respect to $\mathbf{M}$'s lines. For a vector $\mathbf{u}$, $\mathbf{u}[k]$ refers to its $k^{th}$ component.
For a given integer $m$, we note $\mathbf{I}_m$ the identity matrix of size $m \times m$.
Let $\mathcal{E}$ be a euclidean space ($\mathcal{E}$ can be $\mathbb{R}^2$ or $\mathbb{R}^3$ if we consider spatial or spatio-temporal data respectively).
We note  $\mathcal{U} = (\mathbf{u}_k)_{1\leq k\leq p}$  a set of vectors in $\mathcal{E}$. In this paper, we only consider the case  $\mathcal{E} = \mathbb{R}^2$; $\mathcal{U}$ will be a set of positions in a plan. 
 
\section{The PSFs field}
\label{dim_reduc}
\subsection{The observation model}

We assume that we have an image $I$, which contains $p$ unresolved objects such as stars, which can be used to estimate the PSFs field. 
Noting $\mathbf{y}_k$ one of these $p$ objects at spatial 
position $\mathbf{u}_k$, $\mathbf{y}_k$  is therefore a small patch of $I$ with  $n_y$ pixels, 
around the spatial position $\mathbf{u}_k$. We will write $\mathbf{y}_k$ as a 1D vector.
The 
relation between the "true" PSF $\mathbf{x}_k$ and the noisy $\mathbf{y}_k$ observation is
\begin{equation}
\label{eq_patch_psf}
\mathbf{y}_k = \mathbf{M}_k  \mathbf{x}_k + \mathbf{n}_k
\end{equation}
where $\mathbf{M}_k$ is a linear operator and $\mathbf{n}_k$ is a noise that we assume to be Gaussian and white. 
We will consider two kinds of operators in this paper: the first one is the simple case where $\mathbf{M}_k= \mathbf{I}_{n_x}$ and 
we have the number of pixels $n_x$ in $\mathbf{x}_k$ is equal to $n_y$, and the second one is a shift+downsampling degradation operator and
 $n_x =  m_d^2 n_y$, where $m_d$ is the downsampling factor in lines and columns, with $m_d\geq 1$.
 
Noting $\mathbf{Y} = [\mathbf{y}_1\dots\mathbf{y}_p]$ the matrix of $n_y$ lines and $p$ columns of all observed patches, 
 $\mathbf{X} = [\mathbf{x}_1\dots\mathbf{x}_p]$ the matrix $n_x \times p$  of all unknown PSFs,  we can  rewrite Eq.~\ref{eq_patch_psf} as
 \begin{equation}
\label{eq_psf}
\mathbf{Y} = \mathcal{F}  ( \mathbf{X} )  + \mathbf{N}
\end{equation}
 where $\mathcal{F}  ( \mathbf{X} ) =[\mathbf{M}_1\mathbf{x}_1,\dots,\mathbf{M}_p  \mathbf{x}_p] $.
 
 This rewriting is useful because, as we discuss in the following, the different 
 PSFs $\mathbf{x_k}$ are not independent, which means that the problems of Eq.~\ref{eq_patch_psf}
 should not be solved independently for each $k$.  In other terms, the vectors $ (\mathbf{x}_k)_{1\leq k\leq p}$ belong to 
 a specific unknown manifold that needs to be learned by using the data globally.
 
\subsection{The data model}
Let $\boldsymbol{\Omega}$ be a $r$ dimensional subspace of $\mathbb{R}^{n_x}$ embedding the PSFs field. We assume that there exists a continuous function $f: \mathcal{E} \mapsto \boldsymbol{\Omega}$, so that $f(\mathbf{u}_k) = \mathbf{x}_k, \; \forall k \in \llbracket 1,p \rrbracket$.
The regularity of $f$ translates the correlation of the data in space (and time).

Let $(\mathbf{s}_i)_{1\leq i\leq r}$ be a basis of $\boldsymbol{\Omega}$. By definition, we can write each $\mathbf{x}_k$ as a linear combination of the $\mathbf{s}_i$, $\mathbf{x}_k = \sum_{i=1}^r a_{ik}\mathbf{s}_i$, $k=1 \dots p$, or equivalently
\begin{equation}
\label{eq_eigenPSF}
\mathbf{X} =    \mathbf{S}   \mathbf{A} 
\end{equation}
where $\mathbf{S} = [\mathbf{s}_1,\dots,\mathbf{s}_r]$ and $\mathbf{A}$ is a $r \times p$  matrix
containing the coefficients $\mathbf{A}[:,k]$ of the vectors $\mathbf{x}_k$ ($k=1 \dots p$) in the dictionary $\mathbf{S}$.
Each column of the matrix $\mathbf{S}$, that we also refer to as an atom, can be seen as an {\em eigen PSF}, i.e. a given PSF's feature distributed across the field.

\subsection{The inverse problem}
\label{sect_inv_pb}
We need therefore to minimize $ \|\mathbf{Y} - \mathcal{F} (  \mathbf{X} )\|_F^2$,  which is an ill posed problem due to both the noise and the operator $\mathcal{F}$, $\|.\|_F$ denoting the Frobenius norm of a matrix.  There are several constraints  that may be interesting to use in order to properly regularize this inverse problem:
\begin{itemize}
\item{\underline{positivity constraint}:} the PSF $\mathbf{x}_k$ should be positive;
\item{\underline{low rank constraint}:} as described above, we can assume that $\mathbf{x}_k = \sum_{i=1}^r a_{ik}\mathbf{s}_i$, which means that we 
can instead minimize  
\begin{equation}
\min_{ \mathbf{A}, \mathbf{S}}  \|\mathbf{Y} - \mathcal{F}(\mathbf{S}\mathbf{A})\|_F^2;
 \end{equation}
we assume that $r \ll \min(n,p)$; this dimension reduction has the advantage  that there are much less unknown to find, leading to more robustness, but the problem is now that the cost function is not convex anymore;
\item{\underline{smoothness constraint}:} we can assume that the vectors $\mathbf{x}_k$ are structured;  the low rank constraint does not necessarily impose $\mathbf{x}_k$ to be smooth or piece-wise smooth; adding an additional constraint on $\mathbf{S}$ atoms, such as a sparsity constraint, allows to capture spatial correlations within the PSFs themselves; an additional dictionary $\mathbf{\Phi}_s$ can therefore be introduced which is assumed to give a sparse representation of the vectors $\mathbf{s}_k$;
\item{\underline{proximity constraint}:}  we can assume that a given $\mathbf{x}_k$ at a position $\mathbf{u}_k$ is very close to another PSF 
$\mathbf{x}_{k^{'}}$ 
at position
$ \mathbf{u}_{k^{'}}$ if the distance between $\mathbf{u}_k$ and $ \mathbf{u}_{k^{'}}$ is small; this means that the field $f$ must be regular;
this regularity can be forced by adding constraints on the lines of the matrix $ \mathbf{A}$;  indeed, the $p$ values relative to a line $\mathbf{A}[i,:]$ 
correspond to the contribution of the $i$th {\em eigen PSF} to locations relative to the spatial positions  $\mathcal{U}$. 
\end{itemize}
We show in section~\ref{sec_rca} how these four constraints can be jointly used to derive the solution.
Let first review existing methods susceptible to solve this problem. 

\section{Related work} 
In all this section, $\mathcal{Y}$ refers to the observed data set $(\mathbf{y}_k)_{1\leq k\leq p}$. In the first part, the aforementioned degradation operator $\mathcal{F}$ is simply the identity. Therefore we review some dimension reduction methods. In the second part $\mathcal{F}$ is a shifting and downsampling operator; we present a PSF modeling software dealing with this more constraining setting. 

\subsection{Dimension reduction} 
The principal components analysis is certainly one of the most popular mathematical procedure in multivariate data analysis and especially, dimension reduction. In our case, we want to represent $\mathcal{ Y}$'s elements using $r$ vectors, with $r \leq \max(p,n_y)$. A PCA gives an orthonormal family of $r$ vectors in $\mathbb{R}^{n_y}$ so that the total variance of $\mathcal{Y}$ along these vectors directions is maximized. By definition, the PCA looks for redundant features over the whole data set. Therefore, in general, the principal components neither capture localized features (in sense of $\mathcal{E}$) nor have a simple physical interpretation.

In \cite{reg_pca}, a "regularized" PCA is proposed to address this shortcoming for spatial data analysis in atmospheric and earth science. Indeed, as a PCA, the method solves the following problem,
\begin{equation}
\underset{\mathbf{A}}\min \;\|\mathbf{Y}-\mathbf{Y}\mathbf{A}^T\mathbf{A}\|_F^2, \; \s.t. \; \mathbf{A}\mathbf{A}^T = \mathbf{I}_r, 
\end{equation}  
for some chosen small $r$. Moreover, it jointly imposes a sparsity constraint and a smoothing penalties with respect to the space $\mathcal{E}$, on the matrix $\mathbf{A}$ lines. This way, with the right balance between those two penalties, one favors the extraction of localized spatial features, making the interpretation of the optimal $\mathbf{A}$ easy. Yet, there is no obvious way of setting the sparsity and smoothness parameters, which are crucial; moreover, unless the data actually contain spatially localized and non-overlapping features, the coupled orthogonality and sparsity constraint is likely to yield a biased approximation of the data.    

In the context of remote sensing and multi-channel imaging, two ways of integrating spatial information into PCA are proposed in  \cite{spat_pca}; the set $\mathcal{Y}$ is made of multi-channel pixels. In the first way, the author introduces a weighting matrix indicating the relative importance of each pixel. For instance, the weight of a given pixel can be related to its distance to some location of interest in $\mathcal{E}$. Then, the computation of the covariance matrix of image bands is slightly modified to integrate this weighting. This idea is close to the methodology proposed in \cite{geo_pca}.  As a consequence, one expects to recover spectral features spatially related to some location of interest within the most important "eigen-pixels". Yet, we do not have any specific location of interest in $\mathcal{E}$ and we rather want to recover relevant features across the whole data set. 

 The second approach aims at taking into account the spatial associations and structural properties of the image. To do so, modified versions of the image bands covariance matrices are calculated, with increasing shifts between the bands, up to a predetermined maximum shifting amplitude. These covariance matrices, including the "regular" one, are averaged and the principal components are finally derived. Intuitively, one expects the spectral features present in structured images regions to be strengthened and therefore captured into the principal components. However, we consider a general setting where the data are randomly distributed with respect to $\mathcal{E}$, which makes the shifted covariances matrices ill-defined. 

A review of PCA applications and modifications for spatial data analysis can be found in \cite{pca_spat_analys}.

In case the data lie on or are close to a manifold $\mathcal{M}$ of dimension $r$ embedded in $\mathbb{R}^n$, one can consider using one of the numerous non-linear dimension reduction algorithms published in the manifold learning literature, such as GMRA \cite{gmra}, \cite{gmra_bounds}. The idea is to partition the data in smaller subsets of sample close to each other in the sense of the manifold geometry. From this partionning, the manifold tangent spaces are estimated at subsets locations; estimates are then simply given by the best regressions of these subsets with $r-$dimensional affine subspaces. The method includes some multiresolution considerations that are not relevant to our problem. This procedure provides a dictionary in which each of the original samples need at most $r$ elements to be represented. Moreover, the local processing of the data, which is necessary in this setting because of the manifold curvature, makes this approach somehow compatible with the considered problem. Indeed, by hypothesis, the closer two samples will be in sense of $\mathcal{E}$, the closer they will be in $\mathbb{R}^n$, and the more likely they will fall into the same local cluster.

Another interesting alternative to the PCA can be found in \cite{lee2008}. This construction called "Treelets" extracts features by uncovering correlated subsets of variables across the data samples. It is particularly useful when the sample size is by far smaller than the data dimensionality ($p \ll n_y$), which does not hold in the application we consider in the following.

\subsection{Super-resolution}
\label{data_field_restore}
In this subsection, $\mathcal{F}$ takes the following form:
\begin{equation}
\mathcal{F}(\mathbf{X}) = [\mathbf{M}_1\mathbf{x}_1^{(c)},\dots,\mathbf{M}_p\mathbf{x}_p^{(c)}],
\label{field_downsamp_op}
\end{equation} 
where $\mathbf{M}_i$ is a warping and downsampling matrix.  
Since we consider a set of compact objects images, the only geometric transformation one has to deal with for registration is the images shifts with respect to the finest pixel grid, which can be estimated using the images centroids \cite{fng}. 

To the best of our knowledge, the only method dealing with this specific setting is the one used in the PSF modeling software PSFEx \cite{bertin}. This method solves a problem of the form:
\begin{equation}
\underset{\boldsymbol{\Delta}_S}\min \frac{1}{2}\|\mathbf{Y} - \mathcal{F}((\boldsymbol{\Delta}_S+\mathbf{S}_0)\mathbf{A})\|_F^2+ \lambda\|\boldsymbol{\Delta}_S\|_F^2.
\end{equation}
$\mathbf{S}_0$ is a rough first guess of the model components. Each line of the weight matrix $\mathbf{A}$ is assumed to follow a monomial law of some given field's parameters. The number of components is determined by the maximal degree of the monomials. For instance, let say that we want to model the PSFs variations as a function of their position in the field with monomials with degrees up to 3, then: 
\begin{itemize}
\item one needs 6 components corresponding to the monomials $1, X, X^2, Y, XY \; \text{and}\; Y^2$;
\item assuming that the $i^{th}$ PSF in $\mathbf{Y}$'s columns order is located at $\mathbf{u}_i = (u_{ix},u_{iy})$ then the $i^{th}$ column of $\mathbf{A}$ is given by $\mathbf{a}_i^{(c)} = [1, u_{ix}, u_{ix}^2, u_{iy},u_{ix}u_{iy},u_{iy}^2]^T$ up to a scaling factor. 
\end{itemize}
This method is used for comparisons in the Numerical experiments part.

\section{Resolved Components Analysis}
\label{sec_rca}
\subsection{Matrix factorization}
\label{data_mod_fond}
\begin{figure*}[ht!]
\begin{center}
\includegraphics[scale=0.5]{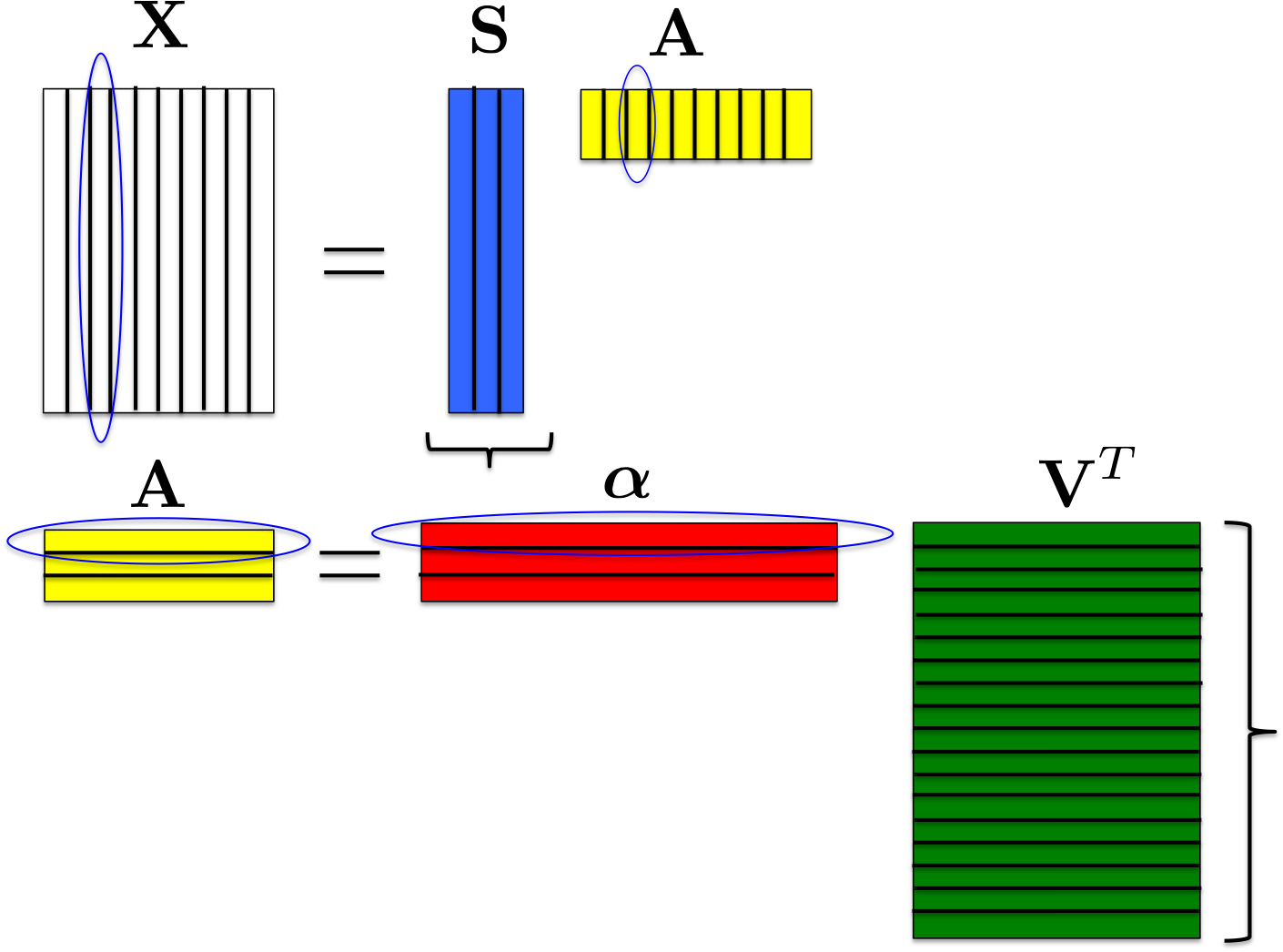}
\end{center}
\caption{Data matrix factorization: the $j^{th}$ sample, which is stored in the $j^{th}$ column of $\mathbf{X}$ is  linear combination of $\mathbf{S}$ columns using $\mathbf{A}$'s $j^{th}$ column coefficients as the weights; similarly, the $j^{th}$ line of $\mathbf{A}$ is a linear combination $\mathbf{V}^T$'s lines, using $\boldsymbol{\alpha}$'s $j^{th}$ line coefficients as the weights.}
\label{facto}
\end{figure*}

We have seen that we can describe the PSFs field $f$ as
\begin{equation}
[f(\mathbf{u}_1),\dots,f(\mathbf{u}_p)] = \mathbf{X} = \mathbf{S}\mathbf{A}. 
\label{decomp} 
\end{equation}
The matrix $\mathbf{S}$ is independent of the spatial location, and 
the $i^{th}$ line of $\mathbf{A}$ gives the contribution of the vector $\mathbf{s}_i$ to each of the samples.
As discussed in section~\ref{sect_inv_pb}, the field's regularity can be taken into account by introducing a structuring of the matrix $\mathbf{A}$. We can write:
\begin{equation}
\label{eq_alpha_decomp}
 \mathbf{A}[i,:]^T = \sum_{l=1}^{N} \alpha_{il} \boldsymbol{\upsilon}_l, i = 1 \dots r,
\end{equation}
where $(\boldsymbol{\upsilon}_l)_{1\leq l\leq N}$ is a set of vectors spanning $\mathbb{R}^p$. Equivalently, we can write $\mathbf{A} = \boldsymbol{\alpha}\mathbf{V}^T$, where $\mathbf{V} = [\boldsymbol{\upsilon}_1,\dots,\boldsymbol{\upsilon}_N]$ and $\boldsymbol{\alpha}$ is a $r \times N$ matrix (see Fig. \ref{facto}).

\subsection*{Physical interpretation}

An interesting way to well interpret $\mathbf{A}$ is to consider the ideal case where the measurements are distributed following a regular grid of locations $\mathcal{U}$. In this case, we can expand the vector $\mathbf{A}[i,:]^T$ using the Discrete Cosine Transform (DCT),
and vectors $\boldsymbol{\upsilon}_i$ in Eq.~\ref{eq_alpha_decomp} are regular cosine atoms, and the column index of the matrix is related the frequency.
Hence, lines relative to high frequencies will be related to quicky varying  PSF components in the field, while lines related to low frequencies will
be related to PSFs stable components.
 In practice, the sampling is not regular and the DCT cannot be used, and $\mathbf{V}$ has to be learned in a way to keep   the {\em harmonic} interpretation valid.
 We want some lines  $\mathbf{A}[i,:]$ to describe stable PSFs components on the FOV, and other to be more related to local behavior.

\subsection{The proximity constraint on $\mathbf{A}$}
\label{prox_cons}
As previously mentioned, we want to account for the PSFs field's regularity by constraining $\mathbf{A}$'s lines. Specifically, we want some lines to determine the distribution of stable features across the PSFs field while we want other lines to be related to more localized features. In order to build this constraint, let first consider the simple case of a one dimensional field of regularly spaced PSFs.
 
\subsubsection{Regularly distributed observations}

We first assume that $\mathcal{E} = \mathbb{R}$.
\begin{figure*}[ht!]
\begin{center}
\begin{tabular}{ccc}
\subfigcapskip = 5pt
\subfigure[Direct domain samples]{\includegraphics[width = 1.\textwidth]{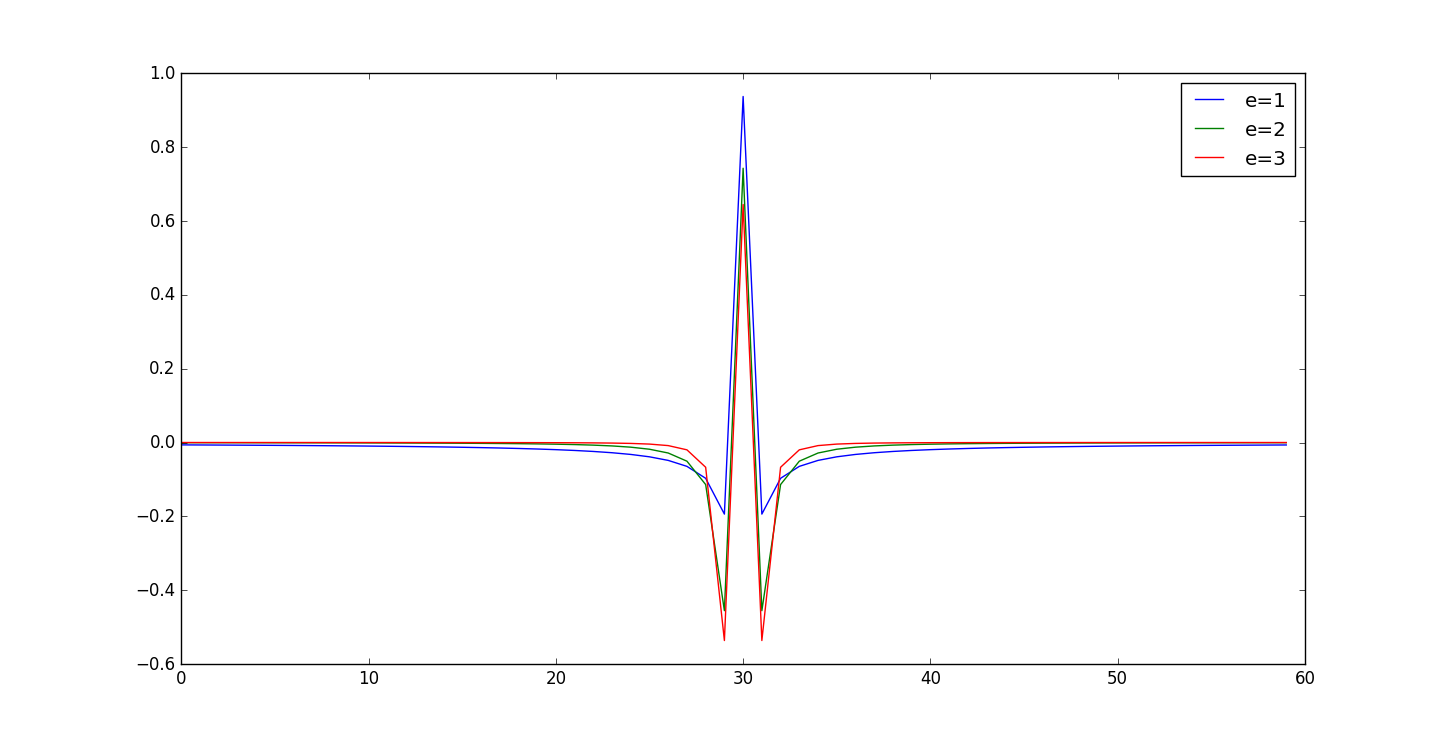}} \\
\label{filters_ex}	
\subfigcapskip = 5pt
\subfigure[Discrete Fourier Transform (DFT) entry-wise moduli]{\includegraphics[width = 0.80\textwidth]{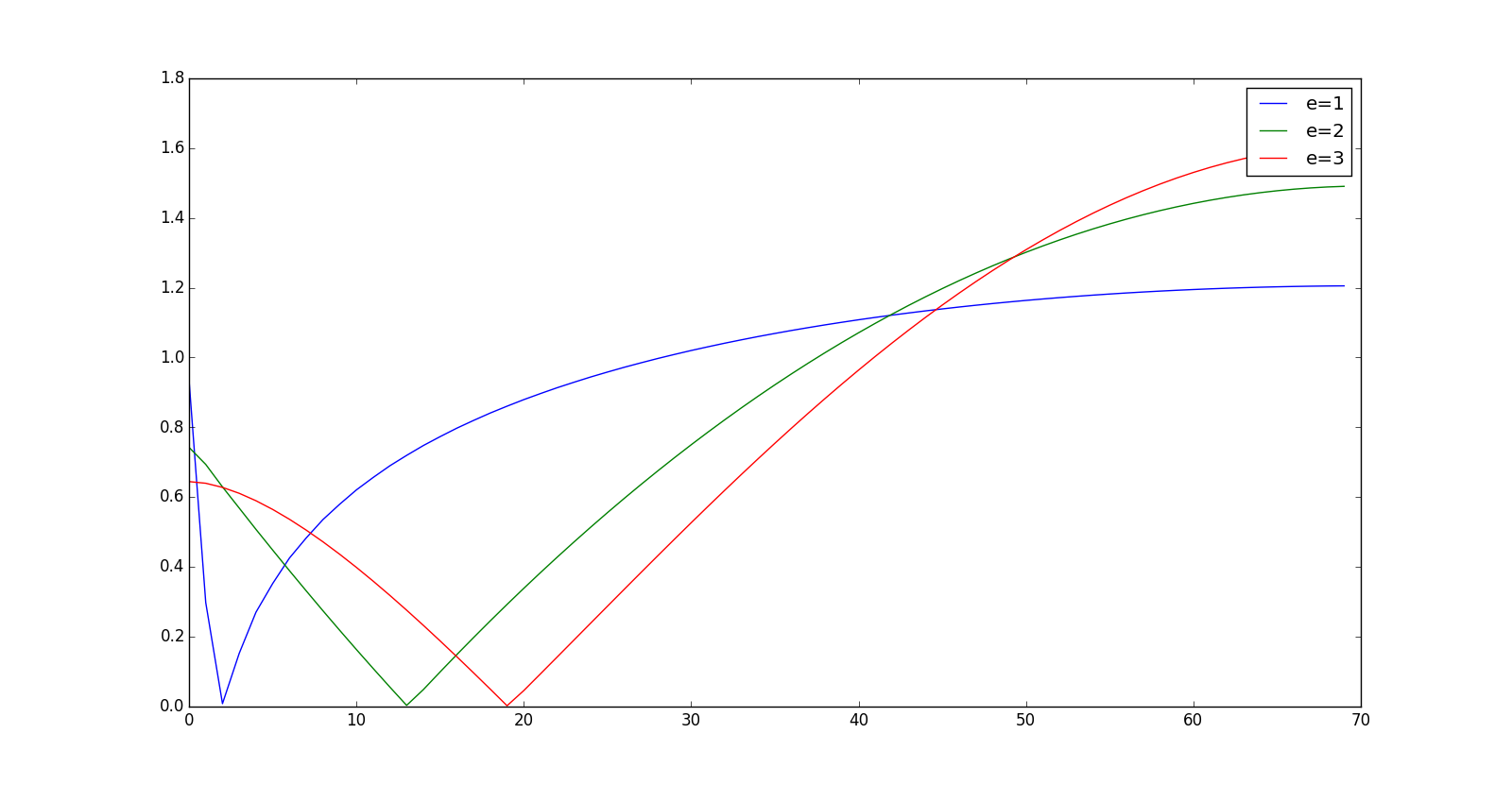}} 
\end{tabular}
\end{center}
\caption{Notch filters examples for different values of the parameter $e$ in Eq. \ref{notch1} and \ref{notch2}. The parameter $a$ is set to 1.}
\label{freq_filters}
\end{figure*}   
We suppose that $p = 2k+1$, for some integer $k$ and
we consider the 1D vector $\boldsymbol{\psi}_{e,a} = (\psi_{i})_{1\leq i \leq p}$ defined as follows:
\begin{eqnarray}
\psi_{i} = \psi_{p-i+1} = -1/|\mathbf{u}_i-\mathbf{u}_{k+1}|^e \;  &  &  \text{if}\; i \neq k+1, \label{notch1} \\
\psi_{i} = \sum_{j=1 \atop j \neq k+1}^p a/|\mathbf{u}_j-\mathbf{u}_{k+1}|^e, &  &  \text{otherwise}
\label{notch2}
\end{eqnarray}
for some positive reals $e$ and $a$. We suppose that $\boldsymbol{\psi}_{e,a}$ is normalized in $l_2$ norm. We refer to this family of signals, parametrized by $e$ and $a$ as "notch filters", in reason of their frequency responses shapes. Some examples can be found in Fig. \ref{freq_filters}.
One can observe that $\mathbf{\psi}_{1,1}$ is essentially a high pass filter. 
As $e$ increases, the notch structure clearly appears, with an increasing notch frequency.  It is clear that, for a vector $ \mathbf{v}$, minimizing the functional
$\Psi_{e,a}(\mathbf{v}) = \|\mathbf{v}\star\boldsymbol{\psi}_{e,a}\|_2^2 $ promotes vectors with spectra concentrated around the notch frequency corresponding to the chosen values of $e$ and $a$. We can directly use this family of filters to constraint $\mathbf{A}$ as follows: we define the functional 
\begin{equation}
\boldsymbol{\Psi}: M_{rp}(\mathbb{R}) \mapsto \mathbb{R}^+\;, \mathbf{A} \rightarrow \sum_{i=1}^r\Psi_{e_{i},a}(\mathbf{A}[i,:]),
\label{notch_mat_penalty}
\end{equation}
where $(e_i)_i$ is a set of reals verifying $0 \leq e_1<e_2<\dots<e_r$ and $a \in [0,2[$. Because the notch frequency increases with $e_i$, minimizing $\boldsymbol{\Psi}$ promotes varying level of smoothness of $\mathbf{A}$'s lines, which is what we wanted to achieve. 
The filter $\boldsymbol{\psi}_{e,a}$ and the functional definitions can be extended to higher dimensions of the space $\mathcal{E}$  by involving a multidimensional convolution \cite{multi_conv}.
Therefore, if the PSFs are distributed over a regular grid with respect to $\mathcal{E}$, one can implement the proximity constraint by solving 
\begin{equation}
\underset{\mathbf{A},\mathbf{S}}\min \frac{1}{2}\|\mathbf{Y} - \mathcal{F}(\mathbf{S}\mathbf{A})\|_F^2 + \lambda \boldsymbol{\Psi}(\mathbf{A}), 
\label{toy_opt} 
\end{equation}
for some positive $\lambda$. Yet, in practical applications, the observations are in general irregularly distributed. In the next section, we propose a slightly different penalty which is usable for arbitrary observations distributions. 

\subsubsection{General setting}
\label{gen_sett_notch}
Let define the functional 
\begin{equation}
\widehat{\Psi}_{e,a}: \mathbb{R}^p \mapsto \mathbb{R}^+\;, \mathbf{v} \rightarrow \sum_{k=1}^p (\sum_{i=1 \atop i \neq k}^p \frac{a v_k-v_i}{\|\mathbf{u}_k-\mathbf{u}_{i}\|_2^e})^2,
\label{notch_penalty_relaxed}
\end{equation}
 where $e$ and $a$ are positive reals. Minimizing $\widehat{\Psi}_{e,a}(\mathbf{v})$ tends to enforce the similarity of close features, with respect to $\mathcal{E}$; in other terms, the more $\|\mathbf{u}_k-\mathbf{u}_{i}\|_2$ is large, the less important is $(a v_k-v_i)$ in $\widehat{\Psi}_{e,a}(\mathbf{v})$ and $e$ somehow determines the radius of similarity. For $e>1$, $\widehat{\Psi}_{e,a} \approx \Psi_{e,a}$ because of the uniform spacing of the values $\mathbf{u}_i$ and the decay of $\frac{1}{\|\mathbf{u}_k-\mathbf{u}_{i}\|_2^e}$, for sufficiently high $p$; we give more details on this approximation in \ref{notch_appd} in the 1D case. However unlike $\Psi_{e,a}$, the functional $\widehat{\Psi}_{e,a}$ is still relevant without the uniform sampling hypothesis and we expect qualitatively the same behavior as $\Psi_{e,a}$ with respect to the frequency domain if the data sampling is sufficiently dense. Therefore we use $\widehat{\Psi}_{e,a}$ instead of $\Psi_{e,a}$ in the functional $\boldsymbol{\Psi}$ of Eq.\ref{toy_opt}. Besides, we use the term frequency even for randomly distributed samples.

\subsubsection{Flexible penalization: the redundant frequencies dictionary $\mathbf{V}$}
 \label{prox_cons_eigen}
The efficiency of the regularization of the problem \ref{toy_opt} relies on a good choice of the parameters 
$e_1,\dots,e_r$ and $a$. Indeed, if the associated notch frequencies does not match with the data set frequency content, the regularization will more or less bias the PSF estimation depending on the Lagrange multiplier $\lambda$. Besides, setting this parameter might be tricky. 
We propose an alternate strategy for constraining $\mathbf{A}$, which leads to the factorization model introduced in Section \ref{data_mod_fond} and still builds over the idea of notch filters. 

For $\mathbf{v} \in \mathbb{R}^p$ we can write
\begin{equation}
\widehat{\Psi}_{e,a}(\mathbf{v}) = \|\mathbf{P}_{e,a}\mathbf{v}\|_2^2,
\end{equation}
where $\mathbf{P}_{e,a}$ is a $p\times p$ matrix defined by
\begin{eqnarray}
\mathbf{P}_{e,a}[i,j] = -\frac{1}{\|\mathbf{u}_i-\mathbf{u}_{j}\|_2^e} \;  &  &  \text{if}\; i \neq j, 
\\
\mathbf{P}_{e,a}[i,i] = \sum_{j=1 \atop j \neq i}^p\frac{a}{\|\mathbf{u}_i-\mathbf{u}_{j}\|_2^e},
\label{notch_mat}
\end{eqnarray}
$(i,j)\in \llbracket 1,p \rrbracket ^2$.
Therefore,
\begin{equation}
\widehat{\Psi}_{e,a}(\mathbf{v}) = \mathbf{v}^T\mathbf{Q}_{e,a}\mathbf{v},
\end{equation}
where $\mathbf{Q}_{e,a} = \mathbf{P}_{e,a}^T\mathbf{P}_{e,a}$ and is symmetric and positive. We consider the singular values decomposition (SVD) of $\mathbf{Q}_{e,a}$: $\mathbf{Q}_{e,a} = \mathbf{V}_{e,a}\mathbf{D}_{e,a}\mathbf{V}_{e,a}^T$.
The diagonal values of $\mathbf{D}_{e,a}$ are sorted in decreasing order. We note $\mathbf{d}_{e,a}$ the vector made of these diagonal values, so that $\mathbf{d}_{e,a}[1]\geq\dots\geq \mathbf{d}_{e,a}[p]\geq 0$. Considering the reduced form $\widehat{\Psi}_{e,a}(\mathbf{v}) = \sum_{i=1}^p \mathbf{d}_{e,a}[i] \langle \mathbf{v},\mathbf{V}_{e,a}[:,i] \rangle^2$, it is clear that minimizing $\widehat{\Psi}_{e,a}(\mathbf{v})$ promotes vectors correlated with $\mathbf{Q}_{e,a}$ last eigenvectors. In the case of regular sampling with respect to $\mathcal{E}$, these eigenvectors are the \underline{harmonics close to the notch frequency of $\boldsymbol{\psi}_{e,a}$}. 
We can rewrite the functional $\boldsymbol{\Psi}$ accordingly: 
\begin{equation}
\boldsymbol{\Psi}(\mathbf{A}) = \sum_{i=1}^r\sum_{j=1}^p \mathbf{d}_{e_i,a}[j] \langle \mathbf{v},\mathbf{V}_{e_i,a}[:,j] \rangle^2. 
\end{equation}
It is clear from this expression that minimizing $\boldsymbol{\Psi}(\mathbf{A})$ enforces the selection of the eigenvectors associated with the lowest eigenvalues in the set $(\mathbf{d}_{e_i,a}[j])_{i,j}$ for describing $\mathbf{A}$'s lines. This can be seen as a sparsity constraint over $\mathbf{A}$'s lines with respect to the atoms $(\mathbf{V}_{e_i,a}[:,j])_{i,j}$; yet, the small subset of atoms which will carry most of the information is somehow predefined through the eigenvalues $(\mathbf{d}_{e_i,a}[j])_{i,j}$. This is unsuitable if the notch filters parameters are poorly selected; on the contrary, one would like to select in a flexible way the atoms which fit the best the data. 
 
Let suppose that we have determined a set of parameters $(e_i,a_i)_{1\leq i\leq r}$ so that the filters $\boldsymbol{\psi}_{e_i,a_i}$ notch frequencies would cover the range of significant frequencies (with respect to $\mathcal{E}$) present in the data. As previously, we note $(\mathbf{V}_{e_i,a_i})_{1 \leq i \leq r}$ the eigenvector's matrices associated with the operators $\widehat{\Psi}_{e_i,a_i}$. We note $\mathbf{V} = [\mathbf{V}_{e_1,a_1},\dots,\mathbf{V}_{e_r,a_r}]$. Considering the preceding remark, we introduce the following problem:
\begin{equation}
\underset{\bm{\alpha},\mathbf{S}}\min \frac{1}{2}\|\mathbf{Y} - \mathcal{F}(\mathbf{S}\boldsymbol{\alpha}\mathbf{V}^T)\|_F^2\; \text{s.t.}\; \|\bm{\alpha}[l,:]\|_0 \leq \eta_l, \; l=1\dots r 
\label{toy_opt_2} 
\end{equation}
Now $\mathbf{A} = \boldsymbol{\alpha}\mathbf{V}^T$. Each line of $\mathbf{A}$ is a sparse linear combination of $\mathbf{V}^T$'s lines, and the "active" atoms are optimally selected according to the data.  
The choice of the parameters $(e_i,a_i)_{1\leq i\leq r}$ and $(\eta_l)_{1\leq i\leq r}$ is discussed in a forthcoming section.  

\subsubsection{A connection with graphs theory}
 \label{graph}
In case $a=1$, the matrix $\mathbf{P}_{e,a}$ is the laplacian of an undirected fully connected and weighted graph with $p$ nodes $1\dots p$, such that the weight of the vertex connecting a node $i$ to a node $j$ is $\frac{1}{\|\mathbf{u}_i-\mathbf{u}_{j}\|_2^e}$\cite{graph_laplace}. As proposed in spectral graph theory \cite{chung1997spectral}, this gives a natural interpretation of $\mathbf{P}_{e,a}$ (and $\mathbf{Q}_{e,a}$) eigenvectors  as harmonic atoms in the graph's geometry. Each line of the matrix $\mathbf{A}$ can be seen as a function defined on a family of graphs determined by the observations locations, so that we enforce the regularity of $\mathbf{A}$'s lines according to the graphs geometry. Our approach is thereby close to the spectral graphs wavelets framework\cite{graph_wav}. However, the graphs wavelets are built on a single graph and a scaling parameter allows one to derive wavelets atoms corresponding to spectral bands of different sizes. In our case, the scales diversity is accounted for by building a dictionary of harmonics corresponding to different graphs. Indeed, as $e$ increases, the weight associated to the most distant nodes (in the sense of $\|\mathbf{u}_i-\mathbf{u}_{j}\|_2$)
becomes negligible, which implies that the corresponding graph laplacian is determined by nearby nodes, yielding "higher" frequencies harmonics. 

\subsection{The smoothness constraint on $\mathbf{S}$}

As previously mentioned, each PSF is a structured image. We can account for this through a sparsity constraint. This has proven effective in multiple frame PSFs super-resolution \cite{fng}.

Since we do not estimate individual PSFs directly, we instead constraint the {\em eigen PSFs} which are $\mathbf{S}$'s columns. 
Specifically, we promote $\mathbf{S}$'s columns sparsity with respect to a chosen dictionary $\boldsymbol{\Phi_s}$. By definition, a typical imaging system's PSF concentrates most of its power in few pixels. Therefore a straightforward choice for $\boldsymbol{\Phi_s}$ is $\mathbf{I}_n$. In other words, we will enforce the sparsity of $\mathbf{S}$'s columns in the pixels domain.

On the other hand, we take $\boldsymbol{\Phi}_s$ as the second generation Starlet forward transform \cite{starlet}, without the coarse scale. The power of sparse prior in wavelet domain for inverse problems being well established, we shall online emphasize the fact that this particular choice of wavelet is particularly suitable for images with nearly isotropic features.

\subsection{Algorithm}
We define the sets $\Omega_1 = \{\boldsymbol{\alpha} \in M_{r,N}(\mathbb{R}) / \|\boldsymbol{\alpha}[l,:]\|_0 \leq \eta_l, \; l=1\dots r \}$ and $\Omega_2 = \{(\mathbf{S},\bm{\alpha}) \in M_{nr}(\mathbb{R})\times M_{r,N}(\mathbb{R}) / \mathbf{S}\bm{\alpha}\mathbf{V}^T\geq_{M_{np}(\mathbb{R})} 0 \}$.
The aforementioned constraints leads us to the following optimization problem:
\begin{equation}
\underset{\boldsymbol{\alpha},\mathbf{S}}\min \frac{1}{2}\|\mathbf{Y} - \mathcal{F}(\mathbf{S}\bm{\alpha}\mathbf{V}^T)\|_F^2 + \sum_{i=1}^r \|\mathbf{w}_i\odot\boldsymbol{\Phi}_s\mathbf{s}_i\|_1\; + \iota_{\Omega_1}(\boldsymbol{\alpha})+ \iota_{\Omega_2}(\mathbf{S},\boldsymbol{\alpha}).   
\label{opt_problem} 
\end{equation} 
where $\odot$ denotes the Hadamard product and $\iota_{\mathcal{C}}$ denotes the indicator function of a set $\mathbf{C}$ (see \ref{cv_analys}). The $\text{l}_1$ term promotes the sparsity of $\mathbf{S}$ columns with respect to $\boldsymbol{\Phi}_s$. The vectors $(\mathbf{w}_i)_i$ weight the sparsity against the other constraints and allow some adaptivity of the penalty, with respect to the uncertainties propagated to each entry of $\mathbf{S}$ \cite{fng}.

The parametric aspects of this method are made clear in the subsequent sections.

The Problem \ref{opt_problem} is globally non-convex because of the coupling between $\mathbf{S}$ and $\bm{\alpha}$ and the $\text{l}_0$ constraint. In particular, the feasible set $\{(\mathbf{S},\bm{\alpha}) \in M_{nr}(\mathbb{R})\times M_{r,N}(\mathbb{R}) / \mathbf{S}\bm{\alpha}\mathbf{V}^T\geq 0 \}$, with $N = rp$ is non-convex. 

Therefore, one can at most expect to find a local minimum. To do so, we consider the following alternating minimization scheme: 
\begin{enumerate}
\item Initialization: $\bm{\alpha}_0 \in \Omega_1$, with $N=rp$, $\mathbf{S}_0 = \underset{\mathbf{S}}\argmin \frac{1}{2}\|\mathbf{Y} - \mathcal{F}(\mathbf{S}\bm{\alpha}_0\mathbf{V}^T)\|_F^2 + \sum_{i=1}^r \|\mathbf{w}_i\odot\boldsymbol{\Phi}_s\mathbf{S}[:,i]\|_1 \; \text{s.t. } \mathbf{S}\bm{\alpha}_0\mathbf{V}^T \geq 0$  
\item For k = 0 \dots $\text{k}_{\text{max}}$: \\
(a) $\bm{\alpha}_{k+1} = \underset{\bm{\alpha}}\argmin \frac{1}{2}\|\mathbf{Y} - \mathcal{F}(\mathbf{S}_k\bm{\alpha}\mathbf{V}^T)\|_F^2\; \text{s.t. } \|\bm{\alpha}[l,:]\|_0 \leq \eta_l, \; l=1\dots r $,

(b) $\mathbf{S}_{k+1} = \underset{\mathbf{S}}\argmin \frac{1}{2}\|\mathbf{Y} - \mathcal{F}(\mathbf{S}\bm{\alpha}_{k+1}\mathbf{V}^T)\|_F^2 + \sum_{i=1}^r \|\mathbf{w}_i\odot\boldsymbol{\Phi}_s\mathbf{S}[:,i]\|_1 \; \text{s.t. } \mathbf{S}\bm{\alpha}_{k+1}\mathbf{V}^T \geq 0$ .
\end{enumerate}
The problem (a) remains non-convex; yet there exists heuristic methods allowing one to approach a local minimum \cite{sous,blum,cartis}. The problem (b) is convex and can be solved efficiently.

 One can note that there is no positivity constraint in the sub-problem (a). This choice is motivated by two facts:
\begin{itemize}
\item the feasible set of (b) is non-empty for any choice of $\bm{\alpha}_{k+1}$;
\item allowing $\bm{\alpha}$ to be outside of the global problem feasible set (for $\mathbf{S}$ fixed) brings some robustness regarding local degenerated solutions.
\end{itemize}

There is an important body of work in the literature on alternate minimization schemes convergence, and in particular in the non-convex and non-smooth setting (see \cite{palm} and the references therein). In the proposed scheme, the analysis is complicated by the asymmetry of the problems (a) and (b).  

We define the function
\begin{equation}
\mathcal{H}(\boldsymbol{\alpha},\mathbf{S}) = \frac{1}{2}\|\mathbf{Y} - \mathcal{F}(\mathbf{S}\bm{\alpha}\mathbf{V}^T)\|_F^2 + \sum_{i=1}^r \|\mathbf{w}_i\odot\boldsymbol{\Phi}_s\mathbf{s}_i\|_1
\label{relaxed_cost}
\end{equation}
and the matrix $\widehat{\mathbf{S}}_k = \underset{\mathbf{S}}{\text{argmin}}\frac{1}{2}\|\mathbf{S}-\mathbf{S}_k\|_2^2 \text{ s.t. } \mathbf{S}\bm{\alpha}_{k+1}\mathbf{V}^T \geq 0$.    
One immediate sufficient condition for the sequence $(\mathcal{H}(\boldsymbol{\alpha}_k,\mathbf{S}_k))_k$ to be decreasing (and thereby convergent) is 
\begin{equation}
\mathcal{H}(\boldsymbol{\alpha}_{k+1},\widehat{\mathbf{S}}_k) \leq \mathcal{H}(\boldsymbol{\alpha}_k,\mathbf{S}_k)
\end{equation}   
which occurs if $(\mathbf{S}_k,\boldsymbol{\alpha}_{k+1})$ stays sufficiently close to $\Omega_2$. Although we do not prove this always holds true, we observe on examples that the matrix $\mathbf{S}_k\boldsymbol{\alpha}_{k+1}\mathbf{V}^T$ in general only has a few and small negative entries for $k\geq 1$. This follows from the adequacy of the dictionary $\mathbf{V}$ for sparsely describing $\mathbf{A}$'s lines.

The complete method is given in Algorithm $\ref{rca_algo}$. The resolution of the minimization sub-problems is detailed in appendices. 


\begin{algorithm*}[!htb]
\caption{Resolved components analysis (RCA)}
\begin{algorithmic}[1]
\label{rca_algo}

\bigskip
\STATE \underline{Parameters estimation and initialization}:  
\\ Harmonic constraint parameters  $ (e_i,a_i)_{1\leq i\leq r} \rightarrow \mathbf{V},\mathbf{A}_0$
\\ Noise level, $\mathbf{A}_0 \rightarrow \mathbf{W}_{0,0}$
\STATE \underline{Alternate minimization}
 \FOR{$k=0$ to $k_{\max}$}
 \FOR{$j=0$ to $j_{\max}$}
 \STATE$\mathbf{S}_k = \underset{\mathbf{S}}\argmin \frac{1}{2}\|\mathbf{Y} - \mathcal{F}(\mathbf{S}\mathbf{A}_k)\|_F^2 + \sum_{i=1}^r \|\mathbf{W}_{k,j}[:,i]\odot\boldsymbol{\Phi}_s\mathbf{S}[:,i]\|_1 \; \text{s.t. } \mathbf{S}\mathbf{A}_k \geq 0$
 \STATE $\text{update: }\mathbf{W}_{k,0},\mathbf{S}_k \rightarrow \text{update}(\mathbf{W}_{k,j+1})$
 \ENDFOR \\
 \STATE $\bm{\alpha}_{k+1} = \underset{\bm{\alpha}}\argmin \frac{1}{2}\|\mathbf{Y} - \mathcal{F}(\mathbf{S}_k\bm{\alpha}\mathbf{V}^T)\|_F^2\; \text{s.t. } \|\bm{\alpha}[l,:]\|_0 \leq \eta_l $
 \STATE $\text{update: Noise level, }\bm{\alpha}_{k+1} \rightarrow \mathbf{W}_{k+1,0}$
 \STATE $\mathbf{A}_{k+1} = \bm{\alpha}_{k+1}\mathbf{V}^T$
 \STATE $\mathbf{A}_{k+1}[i,:] = \mathbf{A}_{k+1}[i,:]/\|\mathbf{A}_{k+1}[i,:]\|_2,\text{ for }i=1\dots r$
 \ENDFOR
 
 \STATE {\bf Return:} $\mathbf{S}_{k_{\max}}$, $\mathbf{A}_{k_{\max}}$.
 
\end{algorithmic}
\end{algorithm*}   

\subsection{Parameters setting}
\label{param_est}
\subsubsection{Components sparsity parameters}
\label{sparse_param}
 We consider the terms of the form $\|\mathbf{w}_{k,j}\odot\boldsymbol{\Phi}_s\mathbf{s}\|_1$, where $k$ is the alternate minimization index and $j$ is the re-weighted $\text{l}_1$ minimization index. We first suppose that $\boldsymbol{\Phi}_s = \mathbf{I}_n$. We decompose $\mathbf{w}_{k,j}$ as:
\begin{equation}
\mathbf{w}_{k,j} = \kappa \bm{\beta}_{k,j}\odot\bm{\lambda}_{k}  
\end{equation} 
  Let consider the minimization problems in $\mathbf{S}$ in Algorithm \ref{rca_algo}. Assuming that we simply minimize the quadratic term using the following steepest descent update rule,
\begin{equation}
\mathbf{S}_{m+1} = \mathbf{S}_{m} + \mu\mathcal{F}^*(\mathbf{Y}-\mathcal{F}(\mathbf{S}_{m}\mathbf{A}_k))\mathbf{A}_k^T,  
\end{equation}    
for a well chosen step size $\mu$, $\mathcal{F}^*$ being the adjoint operator one can estimate the entry-wise standard deviations of the noise which propagates from the observations to the current solution $\mathbf{S}_{m+1}$.
For a given matrix $\mathbf{X}$ in $M_{np}(\mathbb{R})$, we assume that $\mathcal{F}$ takes the following general form $\mathcal{F}(\mathbf{X}) = [\mathbf{M}_1\mathbf{X}[:,1],\dots,\mathbf{M}_p\mathbf{X}[:,p]]$. We define $\mathcal{F}^2$ as:
\begin{equation}
\mathcal{F}^2(\mathbf{X}) = [(\mathbf{M}_1\odot\mathbf{M}_1)\mathbf{X}[:,1],\dots,(\mathbf{M}_p\odot\mathbf{M}_p)\mathbf{X}[:,p]]
\end{equation}
We note $\mathbf{B}$ the observational noise (or model uncertainty) that we assume to gaussian, white and centered. The propagated noise entry-wise standard deviations are given by 
\begin{equation}
\bm{\Sigma}_k = \mu \sqrt{\mathcal{F}^{2*}(\var(\mathbf{B}))(\mathbf{A}_k^T\odot\mathbf{A}_k^T)},
\label{std_1}
\end{equation} 
where $\var()$ returns entry-wise variances and $\mathcal{F}^{2*}$ is the adjoint operator of $\mathcal{F}^{2}$. 
 Now one can proceed to a hypothesis testing on the signal presence in each entry of $\mathbf{S}_{m+1}$ based on $\bm{\Sigma}_k$ \cite{stk5}, and denoise $\mathbf{S}_{m+1}$ accordingly. For instance, we define the noise-free version of $\mathbf{S}_{m+1}$ as follows:
 \begin{equation}
 \hat{\mathbf{S}}_{m+1}[i_1,i_2] = \left\lbrace \begin{matrix} 0, \; \text{if} \; |\mathbf{S}_{m+1}[i_1,i_2]|\leq \kappa \bm{\Sigma}_k[i_1,i_2] \\
	\frac{\mathbf{S}_{m+1}[i_1,i_2]}{|\mathbf{S}_{m+1}[i_1,i_2]|}(|\mathbf{S}_{m+1}[i_1,i_2]|-\kappa \bm{\Sigma}_k[i_1,i_2]). \; \text{otherwise};
	\label{soft_den}
	\end{matrix}\right.\
 \end{equation}
where $\kappa$ controls the false detection probability; the noise being gaussian, we typically choose 3 or 4 for $\kappa$.

The sequence $(\hat{\mathbf{S}}_{m})$ converges to a solution of the problem
\begin{equation}
\underset{\mathbf{S}}\argmin \frac{1}{2}\|\mathbf{Y} - \mathcal{F}(\mathbf{S}\bm{\alpha}_k\mathbf{U}^T)\|_F^2 + \sum_{i=1}^r \kappa\|\bm{\lambda}_{k}[:,i]\odot\mathbf{S}[:,i]\|_1,
\end{equation}
for $\bm{\lambda}_{k} = \kappa/\mu \bm{\Sigma}_k$. One can find some material on minimization schemes in Appendice \ref{min_scheme}. This choice yields a noise-free but biased solution because of the thresholding; this is a well-known drawback of $\text{l}_1$ norm based regularizations. The purpose of the vector $\bm{\beta}_{k,j}$ is to mitigate this bias\cite{cand}. $\bm{\beta}_{k,0}$ is a vector with ones at all entries. At the step 6 in Algo \ref{rca_algo}, $\bm{\beta}_{k,j}$ is calculated as follows:
\begin{equation}
\bm{\beta}_{k,j} = \frac{1}{1+ \frac{|\mathbf{S}_k|}{\kappa\bm{\lambda}_{k}}},
\end{equation}    
where all the operations are entry-wise and $|\mathbf{S}_k|$ is the vector made of element-wise absolute values of $\mathbf{S}_k$ entries. Qualitatively, this removes the strongest features from the $\text{l}_1$ norm terms by giving them small weights, which makes the debiasing possible; conversely, the entries dominated by noise get weights close to 1, so that the penalty remains unchanged.

For $\boldsymbol{\Phi}_s \neq \mathbf{I}_n$ we follow the same rational. To set the sparsity in the transform domain according to the noise induced uncertainty, we need to further propagate it (the noise) through the operator $\boldsymbol{\Phi}_s$. Formally, we need to estimate the element-wise standard deviations of $\mu\boldsymbol{\Phi}_s\mathcal{F}^{*}(\mathbf{B})\mathbf{A}_k^T$. Let consider the intermediate random matrix $\mathbf{Y}_F = \mathcal{F}^{*}(\mathbf{B})$. Assuming that
\begin{equation}
\mathcal{F}(\mathcal{F}^{*}(.)) = \lambda \text{Id}(.),
\label{tight_frame}
\end{equation}
$\mathbf{Y}_F$'s lines are statistically independent. Therefore, within a given column of $\mathbf{Y}_F\mathbf{A}_k^T$, the entries are statistically independent from one another. We deduce that the element-wise standard deviations of $\mu\boldsymbol{\Phi}_s\mathcal{F}^{*}(\mathbf{B})\mathbf{A}_k^T$ are given by 
\begin{equation}
\bm{\Sigma}_k = \mu \sqrt{(\boldsymbol{\Phi}_s\odot\boldsymbol{\Phi}_s)\mathcal{F}^{2*}(\var(\mathbf{B}))(\mathbf{A}_k^T\odot\mathbf{A}_k^T)}.
\label{std_2}
\end{equation}
Then $\bm{\lambda}_{k}$ is obtained as previously and $\bm{\beta}_{k,j}$ is calculated as 
\begin{equation}
\bm{\beta}_{k,j} = \frac{1}{1+ \frac{|\boldsymbol{\Phi}_s\mathbf{S}_k|}{\kappa\bm{\lambda}_{k}}}.
\end{equation}  
The property \ref{tight_frame} is approximately true in the case of super-resolution.

\subsubsection{Number of components}
We do not propose a method to choose the number of components $r$. Yet, we observe that because of the sparsity constraint, some lines of the matrix $\alpha_{k+1}$ at the step 8 in Algorithm \ref{rca_algo} are equal to the null vector, when the number of components is overestimated. The corresponding lines in $\mathbf{A}_{k+1}$ and subsequently the corresponding columns in $\mathbf{S}_k$ are simply discarded. This provides an intrinsic mean to select the number of components. Thus in practice, one can choose the initial r as the data set dimensionality from the embedding space point of view, which can be estimated based on a principal component analysis.

\subsubsection{Proximity constraint parameters}
\label{harmonic_param}
In this section, we consider the functionals $\widehat{\Psi}_{e_i,a_i}$ and especially the choice of the parameters $e_i$ and $a_i$. Let assume that we have determine a suitable range for the parameters: $(e_i,a_i) \in \mathcal{S} = [e_{\text{min}},e_{\text{max}}] \times [a_{\text{min}},a_{\text{max}}]$ for $i=1\dots r$.

For a particular $(e,a)$ we consider the matrix $\mathbf{Q}_{e,a}$ and its eigenvectors matrix $\mathbf{V}_{e,a}$ introduced in section \ref{prox_cons_eigen}. As previously stated, we want the weights matrix $\mathbf{A}$ lines to be sparse with respect to $\mathbf{Q}_{e,a}$'s eigenvectors. In order to choose the parameters and initialize the weights matrix, we use the following greedy procedure. We consider a sequence of matrices $(\mathbf{R}_i)_{1\leq i\leq r}$, with $\mathbf{R}_1 = \mathbf{Y}$. For $i\in \llbracket 1,r\rrbracket$ we define 
\begin{equation}
\mathcal{J}_{e,a}(\mathbf{R}_i) = \underset{k \in\llbracket 1,p\rrbracket}\max \|\mathbf{R}_i\mathbf{V}_{e,a}[:,k]\|_2,
\label{max_eigen}
\end{equation} 
and we note $\mathbf{v}_{e,a}^*$ the optimal eigenvector. We choose the $i^{th}$ couple of parameters as:
\begin{equation}
(e_i,ai) = \underset{(e,a) \in \mathcal{S}}\argmax \mathcal{J}_{e,a}(\mathbf{R}_i).
\label{param_choice}
\end{equation}
$\mathbf{A}_0[i,:] = \mathbf{v}_{e_i,a_i}^*$ and $\mathbf{R}_{i+1} = \mathbf{R}_i - \mathbf{R}_i\mathbf{V}_{e,a}\mathbf{V}_{e,a}^T$.


Regarding the set $\mathcal{S}$, we choose the interval $a_{\text{min}}=0$ and $a_{\text{max}}=2$. This range allows the notch structure, assuming that $e_{\text{min}} \geq 0$; for $a<0$, $\mathbf{h}_{e,a}$ behaves as a low pass filter. For $a\geq0$, we observe that $\mathbf{h}_{e,a}$ becomes a notch filter, with a notch frequency close to the null frequency for $a\geq 2$. 
As previously stated, $e$ determines the influence of two samples on one another corresponding coefficients in the matrix $\mathbf{A}$ in the algorithmic process. According to Section \ref{gen_sett_notch}, we set $e_{\text{min}} = 1$. Let consider the graph $\mathcal{G}_e$ introduced in section $\ref{graph}$. The higher is $e$, the lower is $\mathcal{G}_e$ connexity. Considering that we are looking for global features (yet localized in the field frequency domain), the highest possible value of $e$ should guarantee that the graph $\mathcal{G}_e$ is connected. This gives us a practical upper bound for $e$. Once $\mathcal{S}$ is determined, we discretize this set, with a logarithmic step, in such a way to have more samples close to $(e_{\text{min}},a_{\text{min}})$ which correspond to low notch frequencies. We solve approximately Problem \ref{param_choice} by taking the best couple of parameters in the discretized version of $\mathcal{S}$. This step is the most computationally demanding, especially for large data samples.

\subsubsection{Weights matrix sparsity parameters}
The parameters $\eta_l$ are implicitly set by the minimization scheme used at step 8 in \ref{rca_algo}. This is detailed in \ref{min_scheme}.

\section{Numerical experiments}
\label{num_res}
In this section, we present the data used to test the proposed method, the simulation realized and comparisons to other existing methods for both dimensionality reduction and super-resolution aspects.
\subsection{Data}
\label{sim_data}
\begin{figure}
\begin{center}
\includegraphics[scale=0.40]{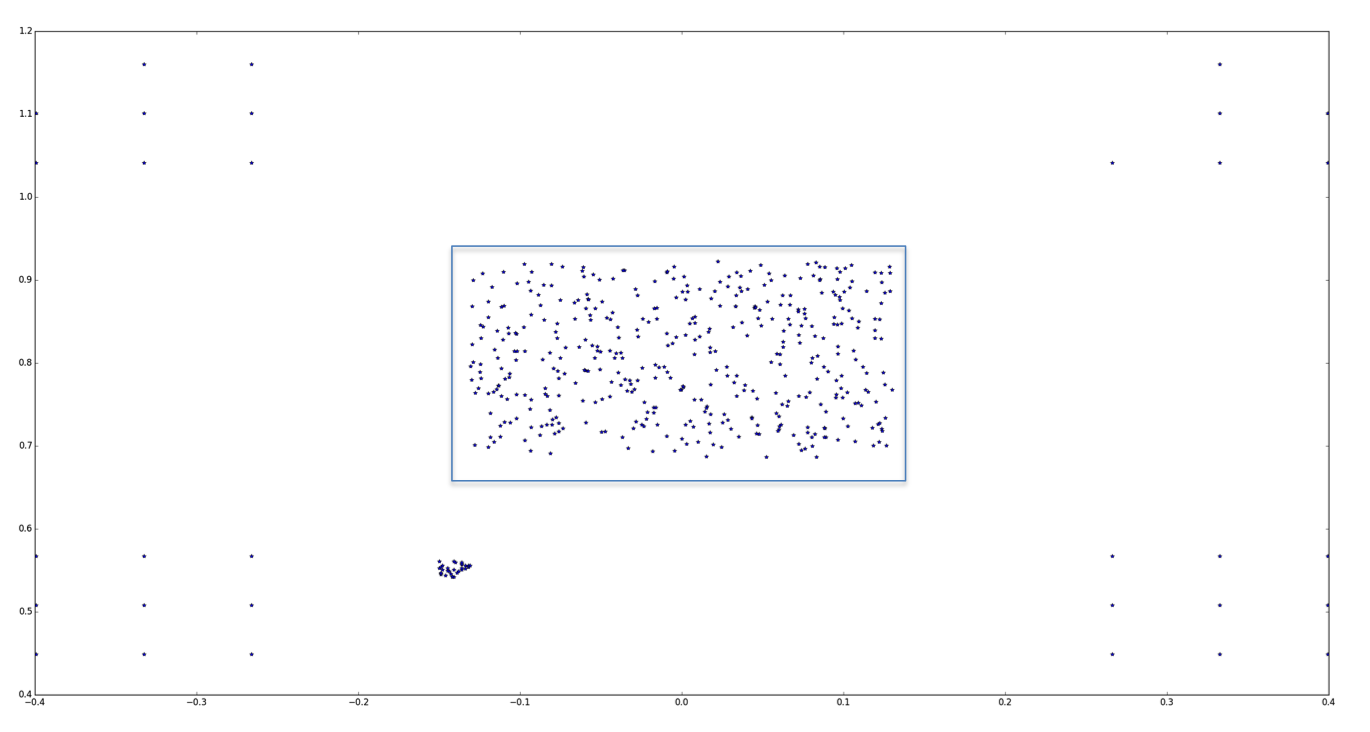}

\end{center}
\caption{Simulated PSFs distribution across the FOV.}
\label{psf_distrib_2}
\end{figure}

The data set consists of simulated optical Euclid PSFs as in \cite{fng}, for a wavelength of $600\mu m$. The PSFs distribution across the field is shown on Fig. \ref{psf_distrib_2}. 
These PSFs account for mirrors polishing imperfections, manufacturing and alignment errors and thermal stability of the telescope.   

\subsection{Simulation}
\label{sim}
We applied different dimension reduction algorithms to a set of 500 PSFs located in the blue box on Fig. \ref{psf_distrib_2}. We applied the algorithms to different observations of the fields, with varying level of white gaussian noise. For a discrete signal $\mathbf{s}$ of length $N$ corrupted with a white gaussian noise $\mathbf{b}$, we define the signal to noise ratio (SNR) as: 
\begin{equation}
\text{SNR} = \frac{\|\mathbf{s}\|_2^2}{N\sigma_\mathbf{b}^2}.
\end{equation} 
\subsection{Quality assessment}
In astronomical surveys, the estimated PSF's shape is particularly important; precisely, one has to be able to capture the PSF anisotropy. 
We recall that for an image $\mathbf{X} = (x_{ij})_{i,j}$, the central moments are defined as 
\begin{equation}
\mu_{p,q}(\mathbf{X}) = \sum_i\sum_j(i-i_c)^p(j-j_c)^q x_{ij} 
\end{equation}
with $(p,q) \in \mathbb{N}^2$, $(i_c,j_c)$ are the image centroid coordinates.
The moments $\mu_{2,0}$ and $\mu_{0,2}$ quantifies the light intensity spreading relatively to the lines $\{(i_c,y), y\in \mathbb{R}\}$ and $\{(x,j_c), x\in \mathbb{R}\}$ respectively. Now we consider the moment $\mu_{1,1}$. We introduce the centered and rotated pixels coordinates $(x_{i,\theta},y_{j,\theta})$ defined by the system of equations
\begin{eqnarray}
x_{i,\theta}\cos(\theta)+y_{j,\theta}\sin(\theta) = i-i_c \\
-x_{i,\theta}\sin(\theta)+y_{j,\theta}\cos(\theta) = j-j_c,
\label{var_change} 
\end{eqnarray}
for some $\theta\in [0,2\pi]$. 
Then we have
\begin{equation}
\mu_{1,1} = \sum_i\sum_j [\frac{\sin(2\theta)}{2}(-x_{i,\theta}^2+y_{j,\theta}^2)+ (2\cos^2(\theta)-1)x_{i,\theta} y_{j,\theta}] x_{ij},
\end{equation}
and in particular, $\mu_{1,1} = \sum_i\sum_j [\frac{1}{2}(-x_{i,\frac{\pi}{4}}^2+y_{j,\frac{\pi}{4}}^2)] x_{ij}$. It becomes clear that $\mu_{1,1}$ quantifies the light intensity spreading with respect to the pixels grid diagonals.

The ellipticity parameters are defined as,
\begin{gather}
e_1(\mathbf{X}) = \frac{\mu_{2,0}(\mathbf{X})-\mu_{0,2}(\mathbf{X})}{\mu_{2,0}(\mathbf{X})+\mu_{0,2}(\mathbf{X})}  \\
e_2(\mathbf{X}) = \frac{2\mu_{1,1}(\mathbf{X})}{\mu_{2,0}(\mathbf{X})+\mu_{0,2}(\mathbf{X})}.
\label{ellipticity}
\end{gather}
We define the vector $\bm{\gamma}(\mathbf{X}) = [e_1(\mathbf{X}),e_2(\mathbf{X})]^T$. This vector characterizes how much $\mathbf{X}$ departs from an isotropic shape and indicates its main direction of orientation. It plays a central theoretical and practical role in weak lensing based dark matter characterization \cite{Dodelson2003292}. 

Another important geometric feature is the so-called PSF size. It has been shown that the size error is a major contributor to the systematics in weak gravitational lensing surveys \cite{paulo}.
We characterize the size of a PSF $\mathbf{X}$ as follows:
\begin{equation}
\text{S}(\mathbf{X}) = (\frac{\sum_i\sum_j((i-i_c)^2+(j-j_c)^2) x_{ij}}{\sum_i\sum_j x_{ij}})^{1/2}. 
\end{equation}
Assuming that a given PSF is a 2D discrete probability distribution, this quantity measures how much this distribution is spread around its mean $[i_c,j_c]^T$.
Let note $(\mathbf{X}_i)_{1\leq i\leq p}$ the set of "original" PSFs and $(\hat{\mathbf{X}}_i)_{1\leq i\leq p}$ the set of corresponding estimated PSFs with one of the compared methods, at a given SNR. The reconstruction quality is accessed through the following quantities:
\begin{itemize}
\item the average error on the ellipticity vector: $\text{E}_{\bm{\gamma}} = \sum_{i=1}^p \|\boldsymbol{\gamma}(\mathbf{X}_i) - \boldsymbol{\gamma}(\hat{\mathbf{X}}_i)\|_2/p$;
\item noting $\boldsymbol{\Gamma} = [\boldsymbol{\gamma}(\mathbf{X}_1) - \boldsymbol{\gamma}(\hat{\mathbf{X}}_1),\dots,\boldsymbol{\gamma}(\mathbf{X}_p) - \boldsymbol{\gamma}(\hat{\mathbf{X}}_p)]$, the dispersion of the errors on the ellipticity vector is measured through the nuclear norm  $\text{B}_{\boldsymbol{\gamma}} = \|\boldsymbol{\Gamma}\|_*$;
\item the average absolute error on the size: $\text{E}_{\text{S}} = \sum_{i=1}^p |\text{S}(\mathbf{X}_i) - \text{S}(\hat{\mathbf{X}}_i)|/p$ in pixels;
\item the dispersion of the errors on the size: $\sigma_{\text{S}} = \text{std}((\text{S}(\mathbf{X}_i) - \text{S}(\hat{\mathbf{X}}_i))_i)$, in pixels.
\end{itemize}

\subsection{Results}
\subsubsection{Dimension reduction}
\label{dim_red}
In this section, we compare RCA to PCA, GMRA and the software PSFEx. We ran a PCA with different number of principal components between 0 and 15. 10 was the value which provided the best results. GMRA input was the data set intrinsic dimension \cite{intrin_dim}, two, since the PSFs only vary as a function of their position in the field; we provided the absolute squared quadratic  error allowed with respect to the observed data based on the observation noise level. For PSFEx, we used 15 components. Finally, RCA used up to 15 components, and effectively, 2 and 4 components respectively for the lowest SNR fields realization. As previously mentioned, we assess the components sparsity's constraint:
\begin{itemize}
\item on the one hand we consider $\boldsymbol{\Phi}_s = \mathbf{I}_n$ which enforces the components sparsity in pixels domain; this is referred to as "RCA" in the plots;
\item on the other hand, we take $\boldsymbol{\Phi}_s$ as the second generation Starlet forward transform \cite{starlet}, without the coarse scale; this is referred to  as "RCA analysis" in the plots.
\end{itemize}

\begin{figure*}[ht!]

\begin{center}
\begin{tabular}{ccc}
\subfigcapskip = 5pt
\subfigure[Average error on the ellipticity vector]{\includegraphics[width = 0.50\textwidth]{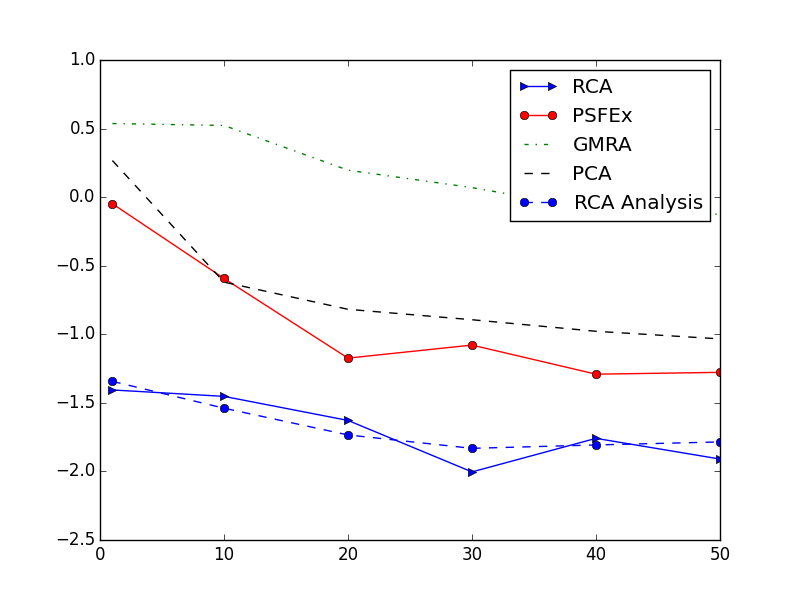}}

\subfigcapskip = 5pt
\subfigure[Dispersion of the ellipticity vector]{\includegraphics[width = 0.50\textwidth]{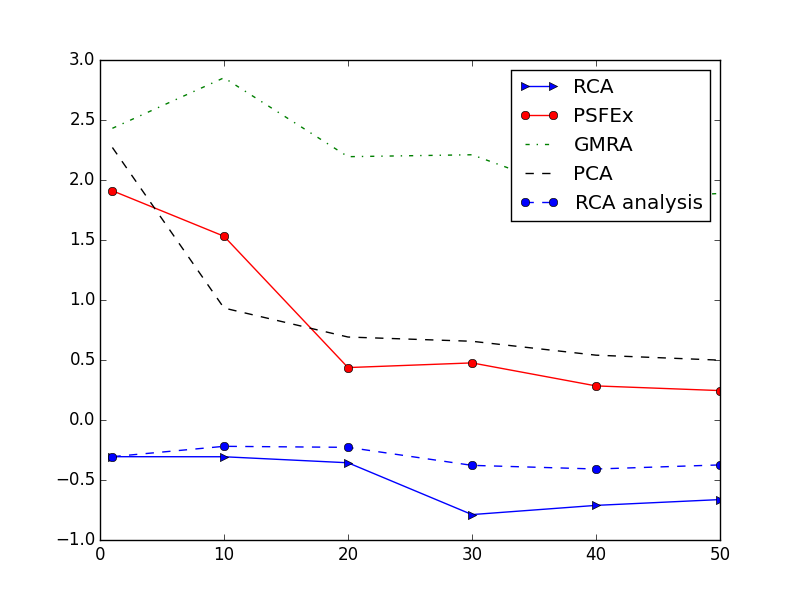}} 
\end{tabular}
\end{center}
\caption{x axis: SNR (see section \ref{sim}); y axis: $\text{log}_{10}(\text{E}_{\bm{\gamma}})$ for the left plot, $\text{log}_{10}(\text{B}_{\bm{\gamma}})$ for the right plot.}
\label{ell_mean_err}
\end{figure*}  

One can see on the left plot in Fig. \ref{ell_mean_err} that the proposed method is at least 10 times more accurate on the ellipticity vector than the other considered methods. Moreover the right plot shows that the accuracy is way more stable. This is true for both choice of the dictionary $\boldsymbol{\Phi}_s$.

\begin{figure*}[ht!]
\begin{center}
\begin{tabular}{ccc}
\subfigcapskip = 5pt
\subfigure[Average absolute error on the size]{\includegraphics[width = 0.50\textwidth]{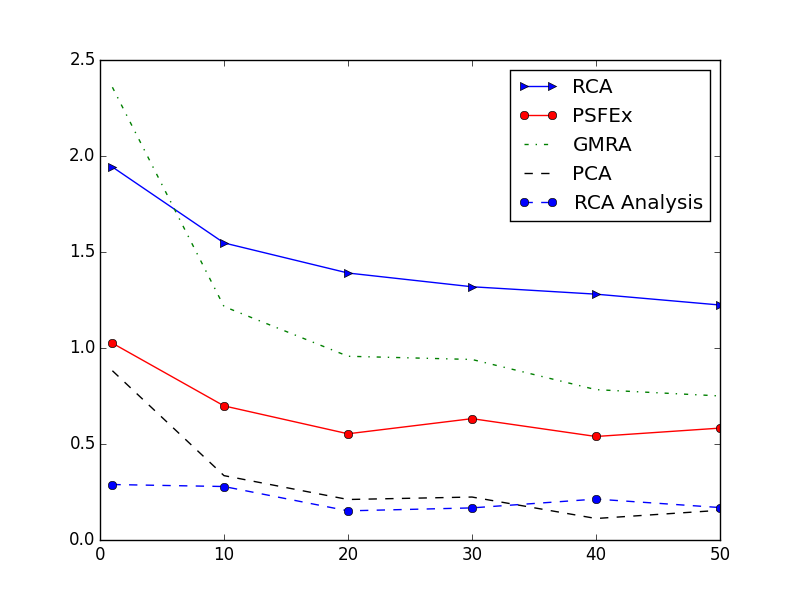}}
	
\subfigcapskip = 5pt
\subfigure[Dispersion of the errors on the size]{\includegraphics[width = 0.50\textwidth]{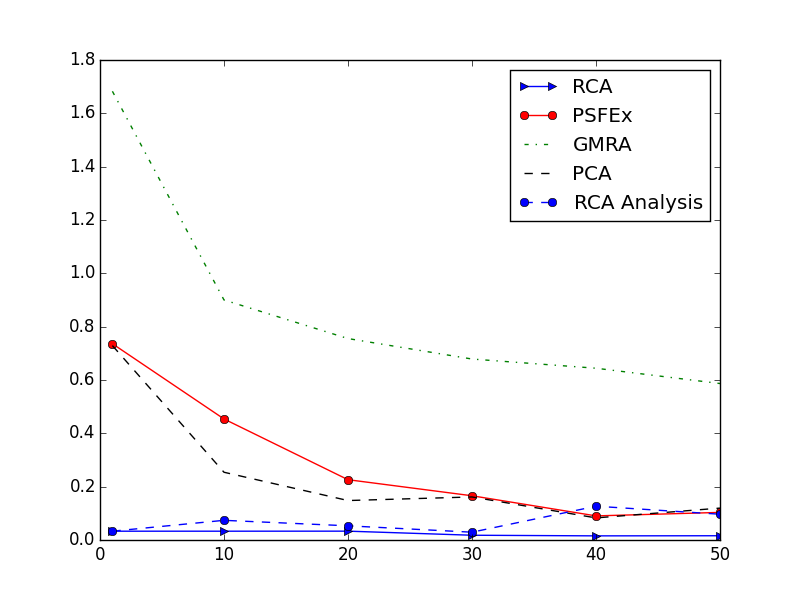}} 
\end{tabular}
\end{center}
\caption{x axis: SNR; y axis: $\text{E}_{\text{S}}$ for the left plot, $\sigma_{\text{S}}$ for the right plot.}
\label{fwhm_mean_err}
\end{figure*} 

Fig. \ref{fwhm_mean_err} shows that the estimated size $\text{S}(\hat{\mathbf{X}}_i)$ is very sensitive to the choice of the dictionary $\boldsymbol{\Phi}_s$. The results are by far more accurate with a sparsity constraint on the components in wavelet domain than in direct domain. 

For a given estimate of the PSF at a given location, the error on the size parameter is more sensitive to errors on the core of the PSF (main lobe and first rings) and less sensitive to errors on the outer part of the PSF than one would expect regarding the error on the ellipticity vector. The error on the outer part of the PSF is essentially related to the observational noise, whereas the error on core of the PSF - which has a high SNR - is more related to the method induced bias. This explains why the PCA performs quite well for this parameter. On the other hand, the bias induced by the sparsity is not only related to the dictionary choice, but also to the underlying data model with respect to the chosen dictionary. 

As previously explained, the components sparsity term is set in such a way to penalize any feature which does not emerge from the propagated noise, which is a source of bias. By using wavelets, we might recover features which are dominated by noise in pixel domain as long as the wavelet filters profile at given scale and direction, matches those features spatial structure. Thus, we expect less error on the reconstructed PSF's core by using wavelets.  
  
We might also consider two distinct ways of using sparsity for the components:
\begin{itemize}
\item we can model each component as $\mathbf{s} = \boldsymbol{\Phi}_s^T\boldsymbol{\alpha}$, with $\boldsymbol{\alpha}$ sparse, which is known in the sparse recovery literature as synthesis prior;
\item we can alternately constraint $\boldsymbol{\Phi}_s\mathbf{s}$ to be sparse. 
\end{itemize}
This priors are equivalent if the dictionary is unitary \cite{elad}. Therefore the pixel domain sparsity constraint can be considered as falling into both framework. However, the two priors are no longer equivalent and potentially yields quite different solutions for overcomplete dictionaries.

We observe in practice that unless the simulated PSFs are strictly sparse with respect to the chosen dictionary - this includes redundant wavelet dictionaries, the synthesis prior yields a bias on the reconstructed PSF size, since the estimated PSFs are sparse linear combinations of atoms which are in general sharper than a typical PSF profile. The analysis prior is somehow weaker and appears to be more suitable for approximately sparse data. 


We do not observe a significant difference between these methods with respect to the mean squared error, except for GMRA which gave noisier reconstructions.
 
\begin{figure*}[ht!]
\begin{center}
\begin{tabular}{ccc}
\subfigcapskip = 2.5pt
\subfigure[Observation 1: center PSF]{\includegraphics[width = 0.18\textwidth]{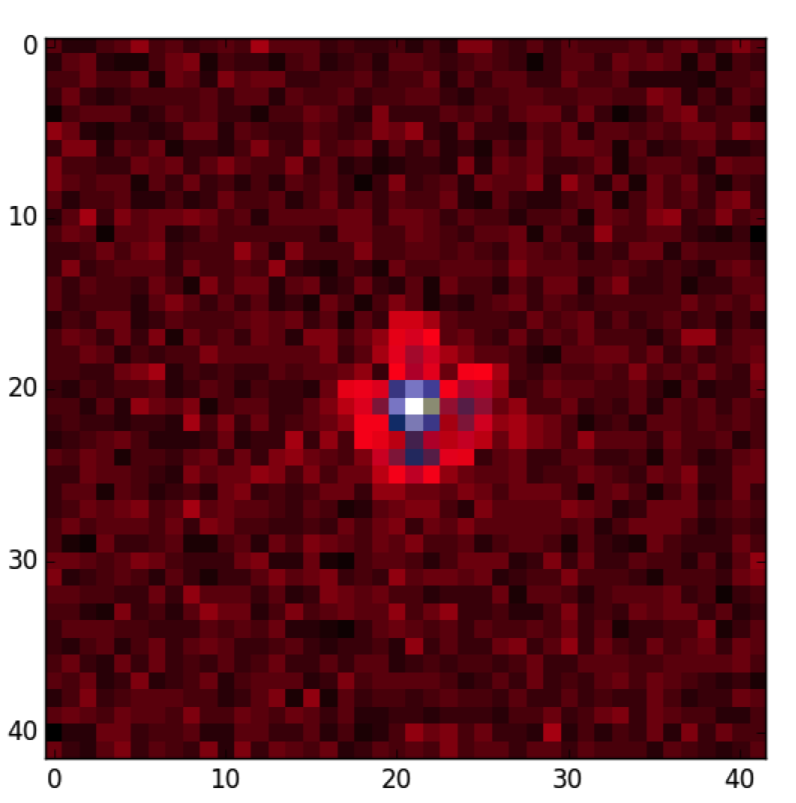}}
\subfigure[Observation 2: center PSF]{\includegraphics[width = 0.18\textwidth]{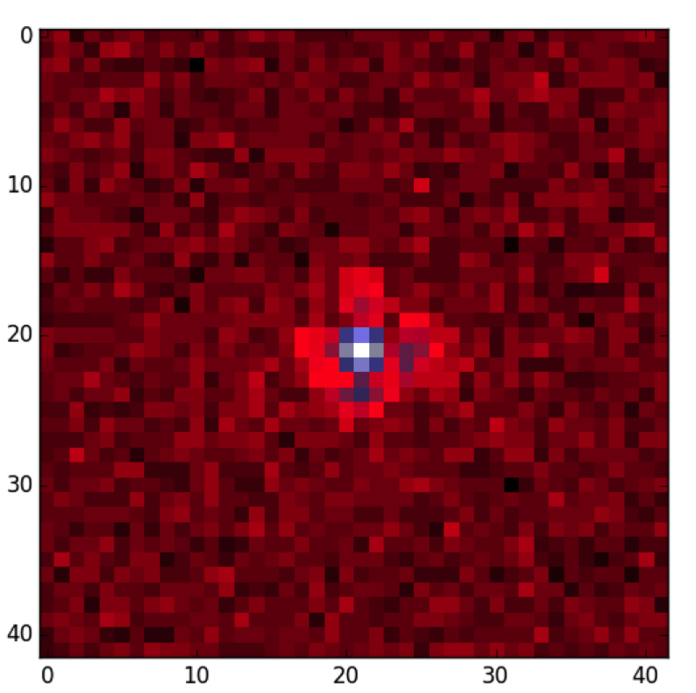}}
\subfigure[Observation 3: corner PSF]{\includegraphics[width = 0.18\textwidth]{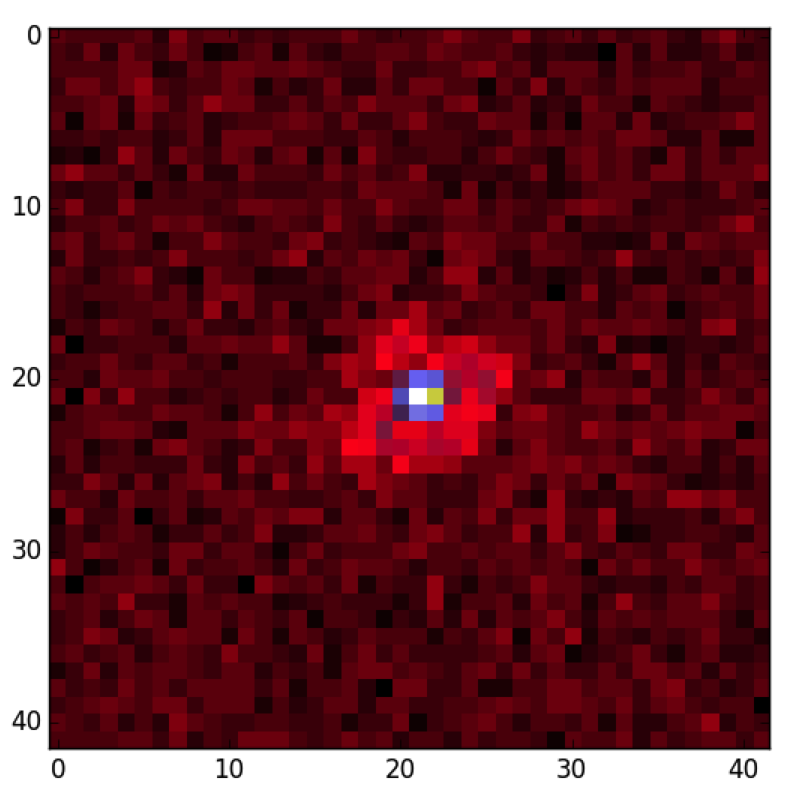}}
\subfigure[Observation 4: corner PSF]{\includegraphics[width = 0.18\textwidth]{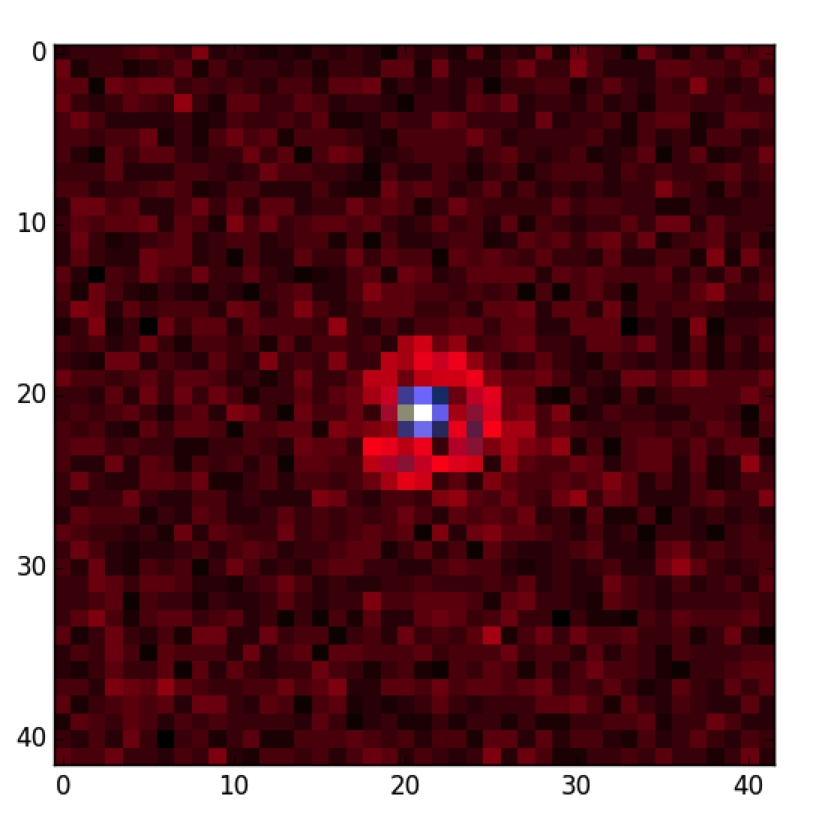}}
\subfigure[Observation 5: "local" PSF]{\includegraphics[width = 0.18\textwidth]{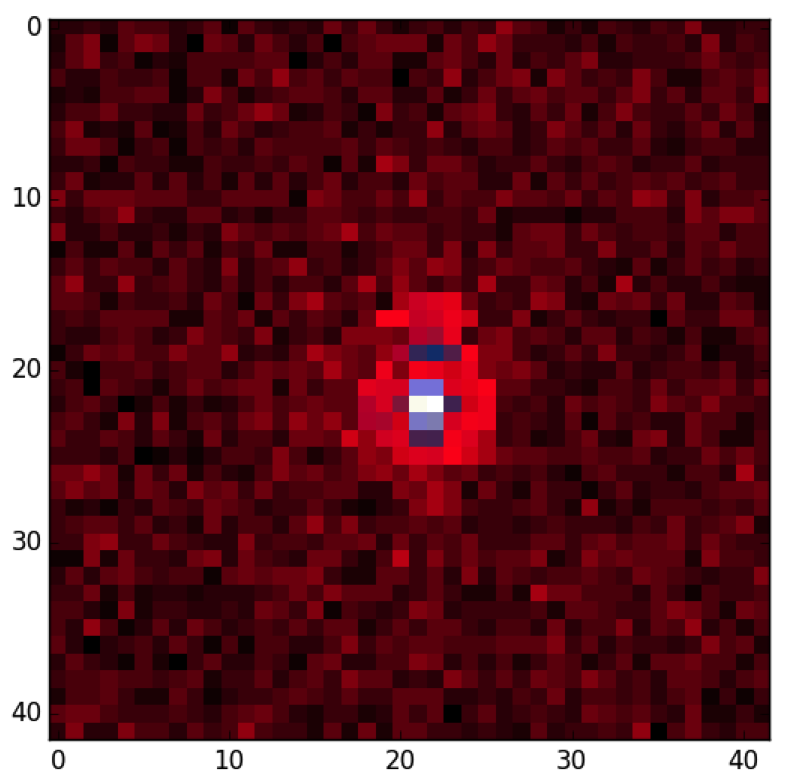}}
\subfigure[Observation 6: "local" PSF]{\includegraphics[width = 0.18\textwidth]{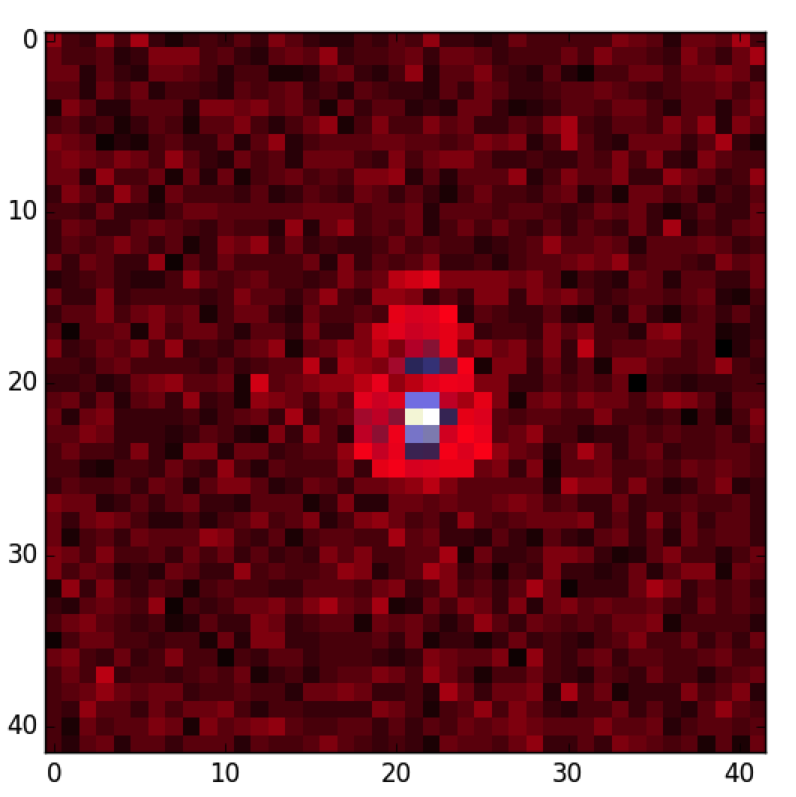}}
\end{tabular}
\end{center}
\caption{Input PSFs at different locations in the FOV for a SNR = 40. The corresponding reconstructed PSFs can be seen in Fig. \ref{rec_hr}}
\label{obs_hr}
\end{figure*}

We applied the aforementioned methods to the PSFs field previously used, with additional 30 corners PSFs and 30 localized PSFs as shown on Fig. \ref{psf_distrib_2} at an SNR of 40. This assess the behavior of the algorithms with respect to spatial clustering and sparse data distribution. 
One can see in Fig. \ref{obs_hr} examples of simulated observed PSFs from different areas in the FOV.   

\bigskip

\begin{figure}
\begin{center}
\includegraphics[scale=0.61]{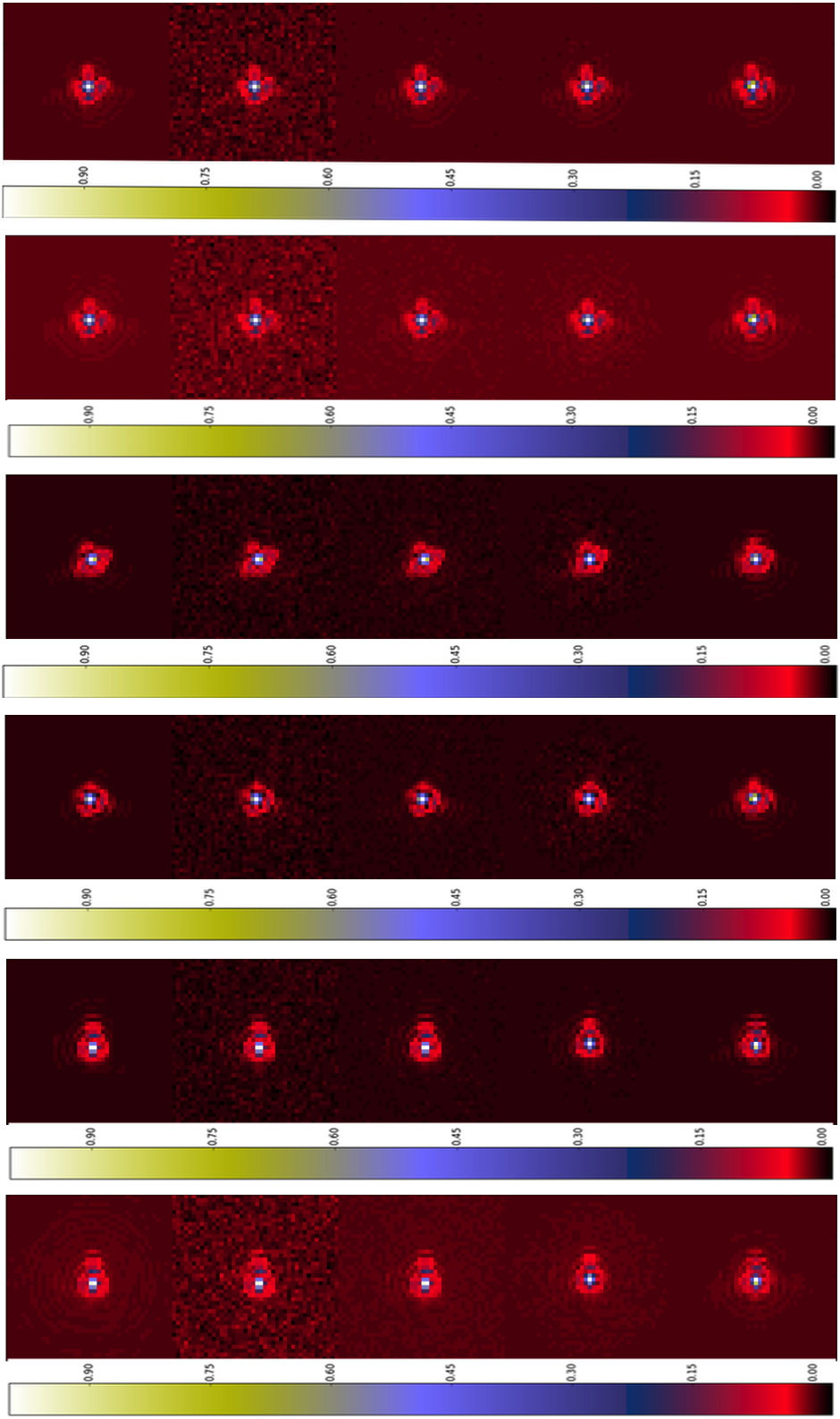}
\end{center}
\caption{PSFs reconstructions: from the left to the right: original, GMRA, PCA, PSFEx, RCA; from the bottom to the top: 2 "local" PSFs reconstructions, 2 corner PSFs reconstructions, 2 center PSFs reconstructions. The observed corresponding PSFs can be seen in Fig. \ref{obs_hr}}
\label{rec_hr}
\end{figure}

For each of these observed PSFs, the reconstructed PSFs for each method are shown in Fig. \ref{rec_hr}. 

%
%
%

One can observe that the proposed method gives noiseless and rather accurate PSFs reconstruction for both the center, the corners and the localized area of the field (see Fig. \ref{psf_distrib_2}). One can also see that we fail to capture accurately the rings pattern in the corners and the localized area. The dictionary $\boldsymbol{\Phi}_s$ considered are not specifically adapted to curve-like structures. The ring patterns varies across the FOV but are locally correlated. Therefore, they can only be recovered where the PSFs are sufficiently dense and numerous, which is the case at the FOV's center. 

PCA and PSFEx yield a significant increase of the SNR in their estimated PSFs at the center and in the localized area. Yet, they fail to do so in the corners because of the lack of correlation for the PCA and local smoothness for PSFEx.

Finally, the poor results obtained with GMRA can be explained by the fact that the underlying manifold sampling is not sufficiently dense for the tangent spaces to be estimated reliably.  

\subsubsection{Super-resolution}

In this section, the data are additionally downsampled to Euclid telescope resolution. PCA and GMRA does not handle the downsampling. Therefore we only consider PSFEx and RCA in this section. For each method, we estimate an upsampled version of each PSF, with a factor 2 in lines and columns; in case of Euclid, this is enough to have a Nyquist frequency greater than half the signal spatial bandwidth \cite{Crop1}.

As previously, RCA Analysis refers to the proposed method, with the dictionary $\boldsymbol{\Phi}_s$ chosen as the second generation Starlet forward transform \cite{starlet}, without the coarse scale; RCA LSQ refers to the proposed method with the dictionary  $\boldsymbol{\Phi}_s$ chosen as the identity matrix, and the weight matrix $\mathbf{A}$ simply calculated as
\begin{equation}
\widehat{\mathbf{A}} = \underset{\mathbf{A}}\argmin \frac{1}{2}\|\mathbf{Y} - \mathcal{F}(\widehat{\mathbf{S}}\mathbf{A})\|_F^2,
\end{equation}
$\widehat{\mathbf{S}}$ being the current estimate of the components matrix.
Among all the methods previously considered for comparison, PSFEx is the only one handling the undersampling. 

\begin{figure*}[ht!]
\begin{center}
\begin{tabular}{ccc}
\subfigcapskip = 5pt
\subfigure[Average error on the ellipticity vector]{\includegraphics[width = 0.50\textwidth]{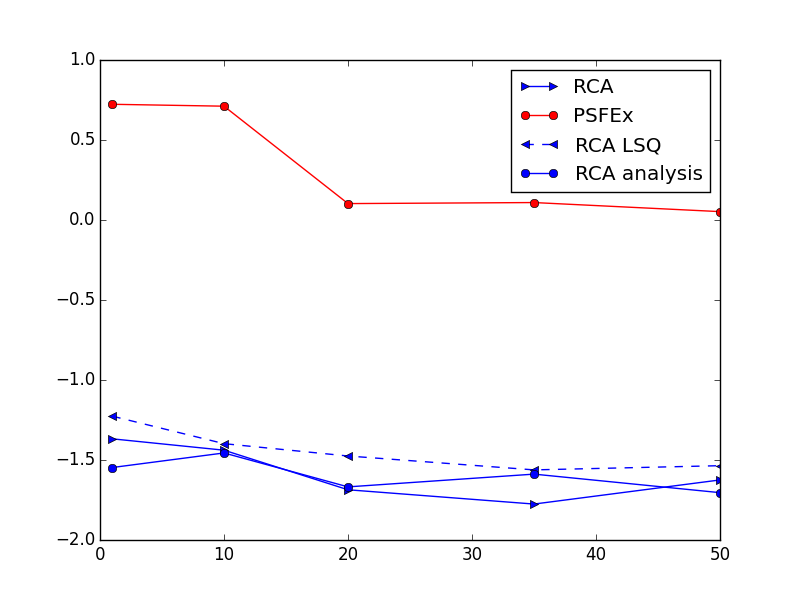}}	
\subfigcapskip = 5pt
\subfigure[Dispersion of the error on the ellipticity vector]{\includegraphics[width = 0.50\textwidth]{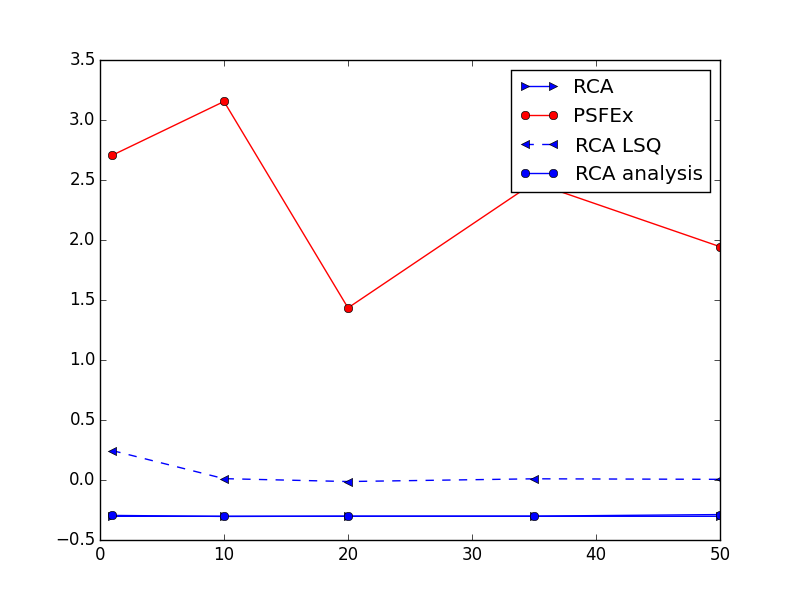}} 
\end{tabular}
\end{center}
\caption{x axis: SNR (see section \ref{sim}); y axis: $\text{log}_{10}(\text{E}_{\bm{\gamma}})$ for the left plot, $\text{log}_{10}(\text{B}_{\bm{\gamma}})$ for the right plot.}
\label{ell_mean_err_2}
\end{figure*}

\begin{figure*}[ht!]
\begin{center}
\begin{tabular}{ccc}
\subfigcapskip = 5pt
\subfigure[Average absolute error on the size]{\includegraphics[width = 0.50\textwidth]{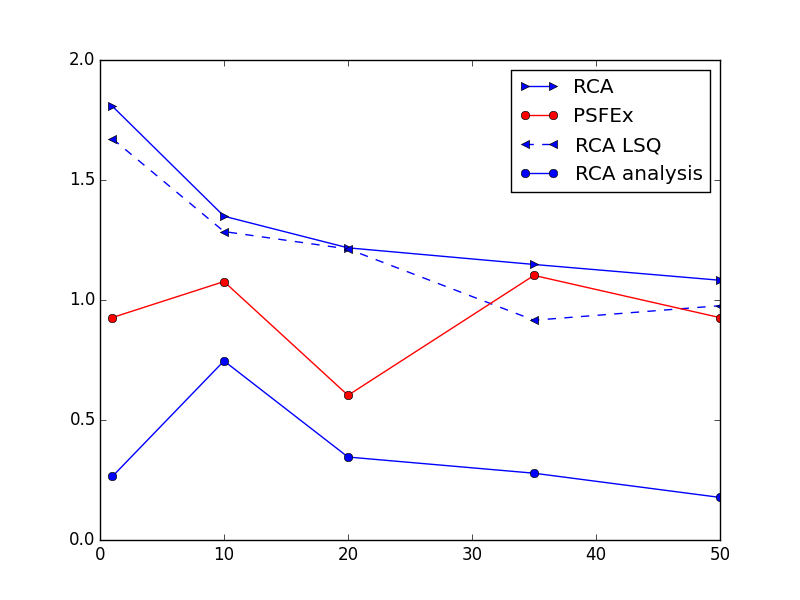}}	
\subfigcapskip = 5pt
\subfigure[Dispersion of the errors on the size]{\includegraphics[width = 0.450\textwidth]{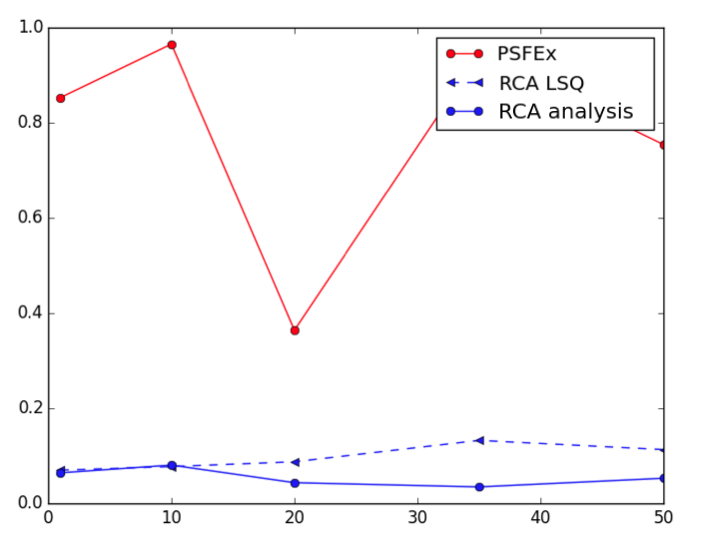}} 
\end{tabular}
\end{center}
\caption{x axis: SNR; y axis: $\text{E}_{\text{S}}$ for the left plot, $\sigma_{\text{S}}$ for the right plot.}
\label{size_mean_err_2}
\end{figure*}

\begin{figure*}[ht!]
\begin{center}
\includegraphics[scale=0.40]{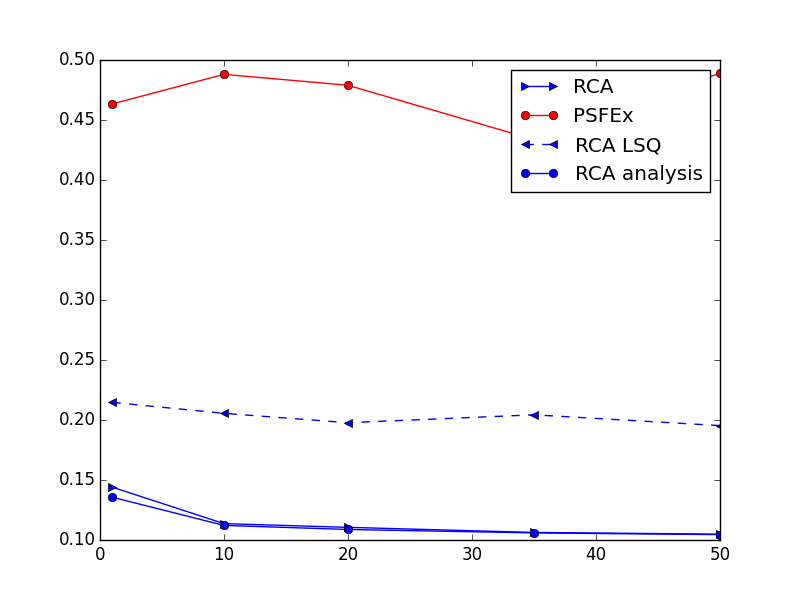}

\end{center}
\caption{Average normalized least square error}
\label{lsq_lr}
\end{figure*} 

As for the dimension reduction experiment, the proposed method with $\boldsymbol{\Phi}_s$ chosen as a wavelet dictionary is at least one order of magnitude more accurate over the shape parameters and the mean square error. Besides, Fig. \ref{lsq_lr} shows that the proximity constraint over the matrix $\mathbf{A}$ allows one to select a significantly better optimum than a simple least square update of $\mathbf{A}$. Indeed, regularizing the weight matrix estimation reinforces the rejection of $\mathcal{F}$'s null space. 

\begin{figure*}[ht!]
\begin{center}
\begin{tabular}{ccc}
\subfigcapskip = 2.5pt
\subfigure[Observation 1: center PSF]{\includegraphics[width = 0.18\textwidth]{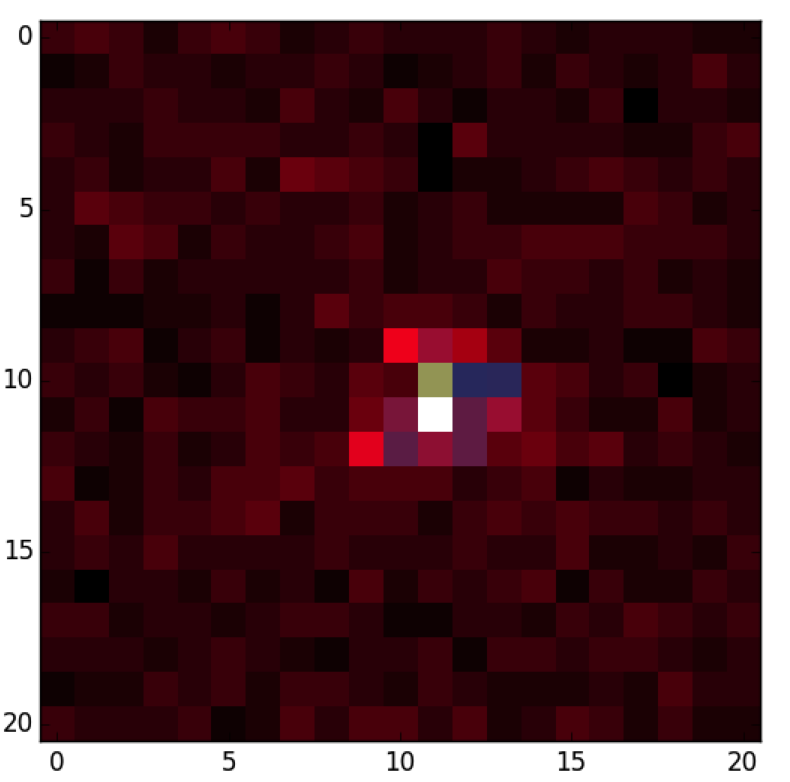}}
\subfigure[Observation 2: center PSF]{\includegraphics[width = 0.18\textwidth]{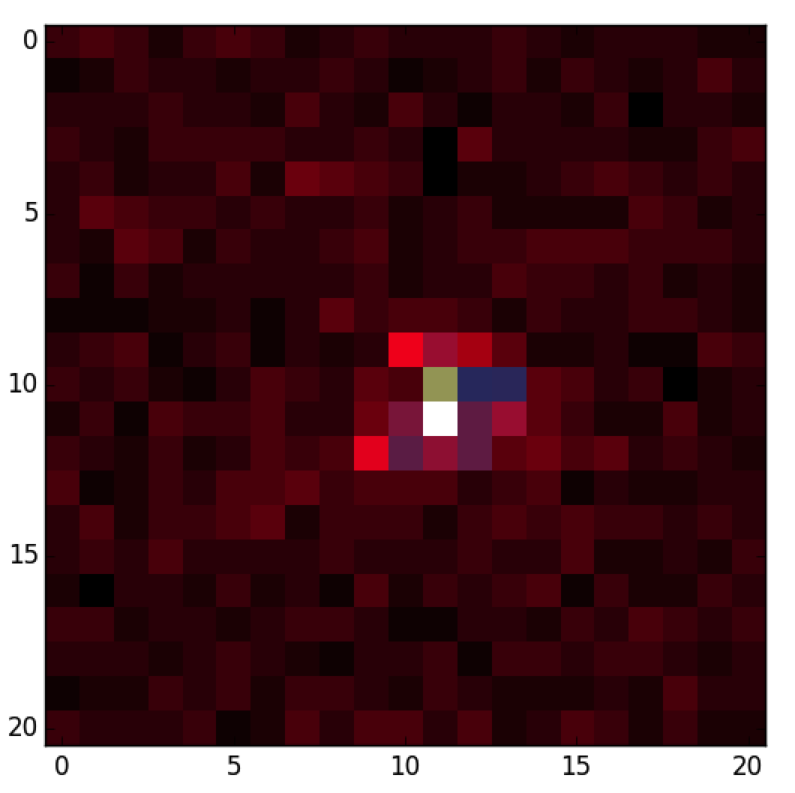}}
\subfigure[Observation 3: corner PSF]{\includegraphics[width = 0.18\textwidth]{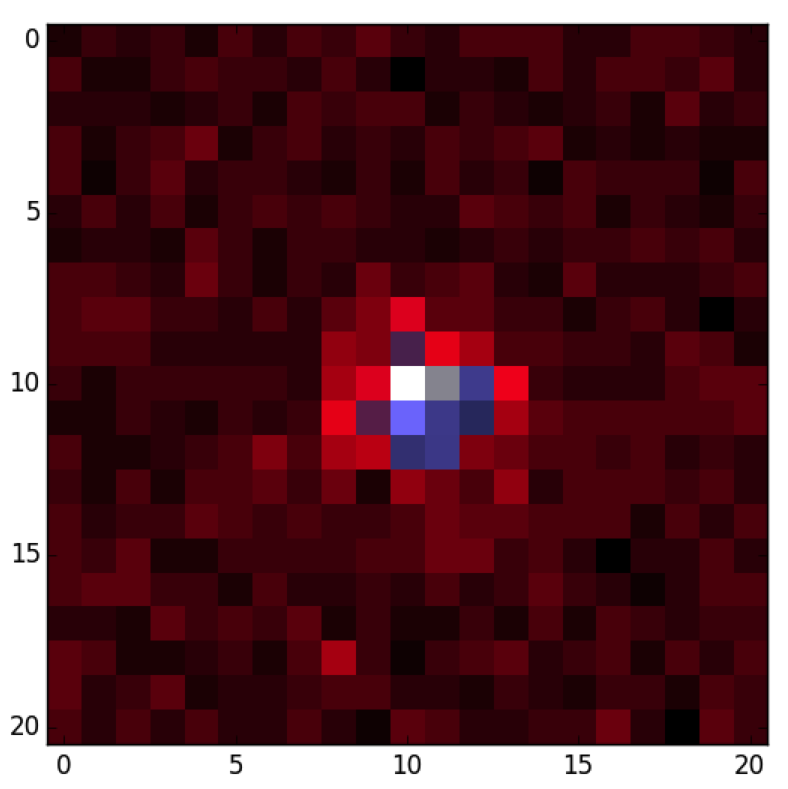}}
\subfigure[Observation 4: corner PSF]{\includegraphics[width = 0.18\textwidth]{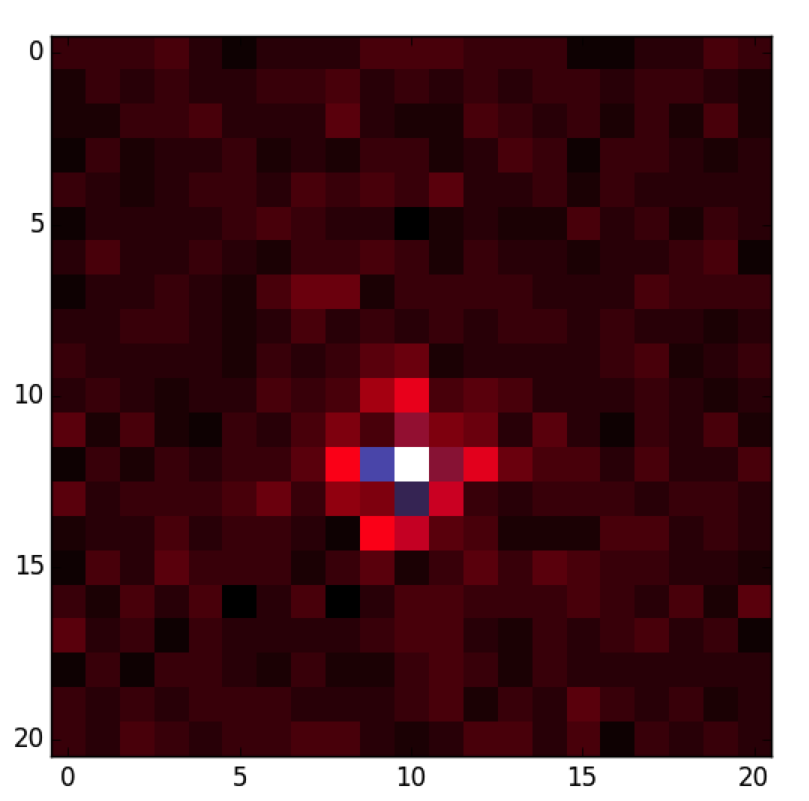}}
\subfigure[Observation 5: "local" PSF]{\includegraphics[width = 0.18\textwidth]{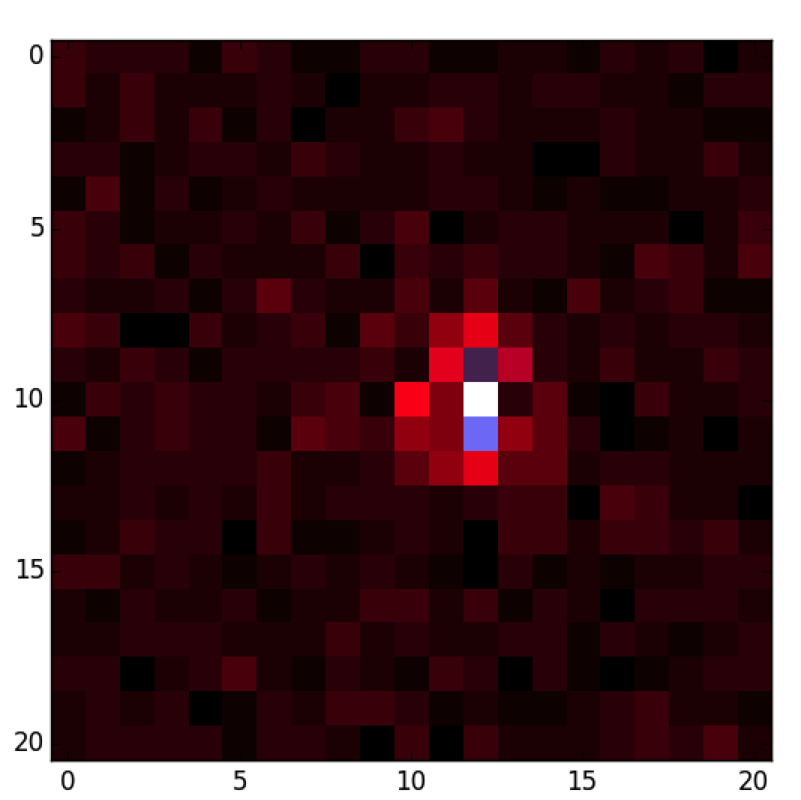}}
\subfigure[Observation 6: "local" PSF]{\includegraphics[width = 0.18\textwidth]{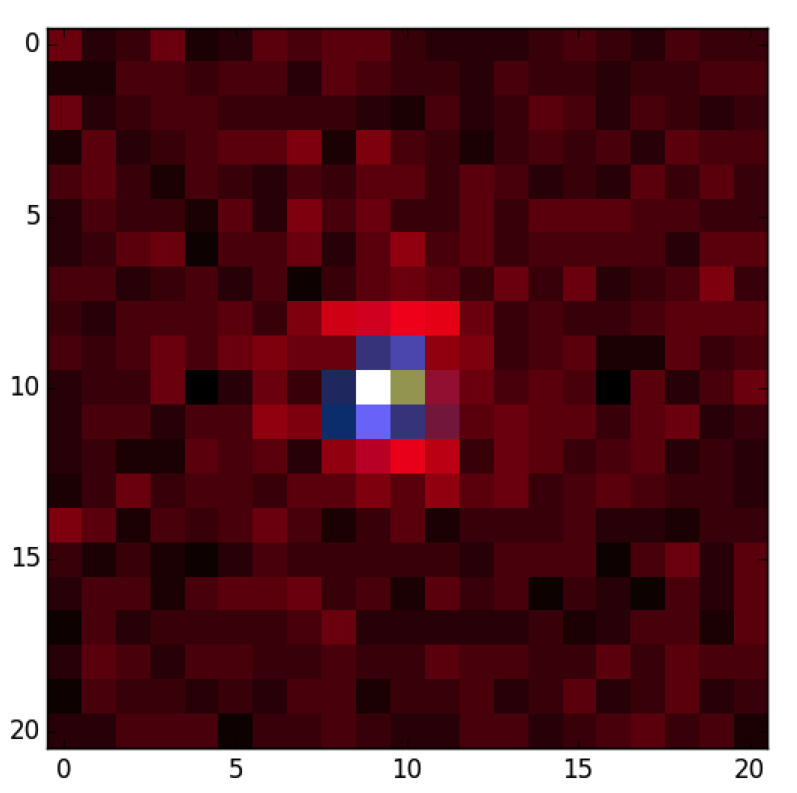}}
\end{tabular}
\end{center}
\caption{Input PSF at different locations in the field for a SNR = 40.}
\label{obs_lr}
\end{figure*} 

As previously, we restored the complete field of Fig. \ref{psf_distrib_2} for a linear SNR of 40, using "RCA Analysis", with undersampled input PSFs as shown in Fig. \ref{obs_lr}. 

%
%
%
%

\begin{figure}
\begin{center}
\includegraphics[scale=0.54]{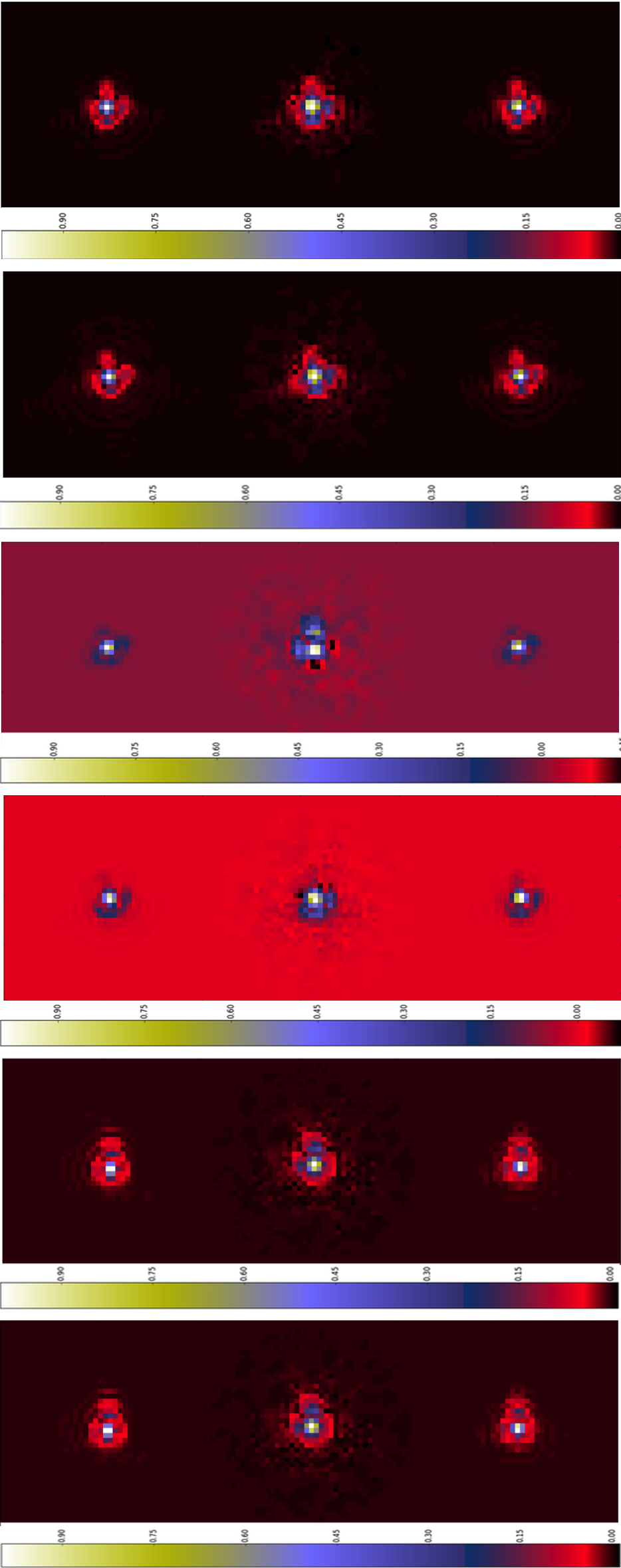}
\end{center}
\caption{PSFs reconstructions: from the left to the right: original, PSFEx, RCA; from the bottom to the top: 2 "local" PSFs reconstructions, 2 corner PSFs reconstructions, 2 center PSFs reconstructions. The observed corresponding PSFs can be seen in Fig. \ref{obs_lr}}
\label{rec_lr}
\end{figure}

The figure \ref{rec_lr} shows consistent results with the dimension reduction experiment. In particular, the corners PSFs restoration is obviously more accurate. 

\section{Reproducible research}
In the spirit of participating in reproducible research, the data and the codes used to generate the plots presented in this paper will be made available at \url{http://www.cosmostat.org/software/rca/}.

\section{Conclusion}
We introduced RCA which is a dimension reduction method for continuous and positive data field which is noise robust and handles undersampled data. As a linear dimension reduction method, RCA computes the input data as linear combinations of few components which are estimated, as well as the linear combination coefficients, through a matrix factorization.

The method was tested over a field of simulated Euclid telescope PSFs.
We show that constraining both the components matrix and  the coefficients matrix using sparsity yield at least one order of magnitude more accurate PSFs restoration than existing methods, with respect to the PSFs shapes parameters. In particular, we show that the analysis formulation of the sparsity constraint over the components is particularly suitable for capturing accurately the PSFs sizes. We also show that constraining the coefficients matrix yields a significantly better identification of the PSFs embedding subspace when the data are undersampled.  

An important extension of RCA for astronomical imaging would be to account for the wavelength dependency of the PSFs. Indeed, an unresolved star image is a linear combination of the PSFs at different wavelengths weighted by the star's spectrum. Hence, RCA can be naturally extended by replacing the matrix $\mathbf{S}$ with a tensor, for which each element would be a polychromatic {\em eigen PSF}.

\section*{Acknowledgements} 
This work is supported by  
the European Community through the grants PHySIS  (contract no. 640174) and  DEDALE  (contract no. 665044) within the H2020 Framework Program. The authors acknowledge the Euclid Collaboration, the European Space Agency and the support of the Centre National d’Etudes Spatiales.
 
\section*{References}
\bibliography{iopart-num}
 
\appendix
\section{Notch filter approximation}
\label{notch_appd}
In this appendix, we explain why the functional $\Psi_{e,a}$ introduced in subsection \ref{prox_cons} can be approximated with the functional $\widehat{\Psi}_{e,a}$. We use the subsection \ref{prox_cons} notations. We consider the 1D case. The samples $(\mathbf{u}_i)_{1\leq i \leq p}$ are uniformly spaced scalar. We assume that $\mathbf{u}_1 < \dots < \mathbf{u}_p$. We note $\Delta = \mathbf{u}_2-\mathbf{u}_1$. Thus,
\begin{eqnarray}
\psi_{i} = \psi_{p-i+1} = \frac{-1}{|k+1-i|^e\Delta^e}\;  &  &  \text{if}\; i \neq k+1,\text{ and} \label{notch1_appd} \\
\psi_{k+1} = 2\sum_{n=1}^k \frac{a}{n^e\Delta^e}.
\label{notch2_appd}
\end{eqnarray}
Using the centered definition of the convolution with a zero boundary condition, for a vector $\mathbf{v} = (v_i)_{1\leq i\leq p}$, the vector $\mathbf{h} = \mathbf{v}\star\boldsymbol{\psi}_{e,a}$ is given by
\begin{equation}
\mathbf{h}[j] = \sum_{i=1}^p v_i \psi_{j+k+1-i},
\label{cent_conv}
\end{equation}
for $j\in \llbracket 1,p \rrbracket$ and with the convention that $\psi_{j+k+1-i} = 0$ if $j+k+1-i<1$ or $j+k+1-i>p$.
Combining Eq.\ref{notch1_appd}, \ref{notch2_appd} and \ref{cent_conv}, we can write 
\begin{equation}
\mathbf{h}[j] = (2\sum_{n=1}^k \frac{a}{n^e\Delta^e})v_j - \sum_{i \in [\max(1,j-k),\atop \min(p,j+k)],  i\neq j} \frac{1}{|j-i|^e\Delta^e}v_i.
\label{conv_exp}
\end{equation}
We recall that $\Psi_{e,a}(\mathbf{v}) = \|\mathbf{h}\|_2^2$. On the other hand, $\widehat{\Psi}_{e,a}(\mathbf{v}) = \|\mathbf{t}_{\mathbf{v}}\|_2^2$, with $\mathbf{t}_{\mathbf{v}}$ defined as
\begin{equation}
\begin{split}
\mathbf{t}_{\mathbf{v}}[j] = (2\sum_{n=1}^{\min(j-1,p-j)}\frac{1}{n^e} + \sum_{n=\atop \min(j-1,p-j)+1}^{\max(j-1,p-j)}\frac{1}{n^e})\frac{a}{\Delta^e}v_j -\sum_{i=1\atop i\neq j}^p\frac{1}{|j-i|^e\Delta^e}v_i \text{ if }j\neq k+1\\
\text{and }\mathbf{t}_{\mathbf{v}}[k+1] = (2\sum_{n=1}^{k}\frac{a}{n^e\Delta^e})v_j-\sum_{i=1\atop i\neq k+1}^p\frac{1}{|k+1-i|^e\Delta^e}v_i.
\label{approx_func}
\end{split}
\end{equation}
Thus, $\mathbf{t}_{\mathbf{v}}[k+1]-\mathbf{h}[k+1] = 0$ and for $j \neq k+1$ 
\begin{equation}
\begin{split}
\mathbf{t}_{\mathbf{v}}[j]-\mathbf{h}[j] = (\sum_{n=k+1}^{\max(j-1,p-j)}\frac{1}{n^e}-\sum_{n=\atop \min(j-1,p-j)+1}^k\frac{1}{n^e})\frac{a}{\Delta^e}v_j\\
 - \sum_{i=1\atop i\neq j}^{\max(1,j-k)}\frac{1}{|j-i|^e\Delta^e}v_i - \sum_{i=\min(p,j+k)\atop i\neq j}^p\frac{1}{|j-i|^e\Delta^e}v_i.
\end{split}
\label{err}
\end{equation}
Given the symmetry of $\boldsymbol{\psi}_{e,a}$ with respect to $k+1$, we focus on the above difference for $j\leq k$. We further assume that $j \neq 1$. Then, Eq.\ref{err} simplifies to
\begin{equation}
\mathbf{t}_{\mathbf{v}}[j]-\mathbf{h}[j] = (\sum_{n=k+1}^{p-j}\frac{1}{n^e}-\sum_{n=j}^{k}\frac{1}{n^e})\frac{a}{\Delta^e}v_j- \frac{1}{(j-1)^e\Delta^e}v_1 - \sum_{n=k}^{p-j} \frac{1}{n^e\Delta^e}v_n.
\label{err_2}
\end{equation}

Now, using the inequalities for $n>1$,
\begin{equation}
\int_{n-1}^n \frac{1}{(t+1)^e}dt \leq \frac{1}{n^e} \leq \int_{n-1}^n\frac{1}{t^e} dt,
\end{equation}
and assuming that $e > 1$, we get the following upper bounding:
\begin{equation}
\begin{split}
|\mathbf{t}_{\mathbf{v}}[j]-\mathbf{h}[j]| \leq \frac{1}{e-1} [\max(|(p-j+1)^{1-e}+(j-1)^{1-e}-(k+1)^{1-e}-k^{1-e}|,\\
|(p-j)^{1-e}+j^{1-e}-(k+1)^{1-e}-k^{1-e}|)a +\frac{e-1}{(j-1)^e}+k^{1-e}-(p-j)^{1-e}]\frac{\|\mathbf{v}\|_{\infty}}{\Delta^e}.  
\end{split}
\end{equation}
We see that the higher is $k$ (we recall that $p=2*k+1$) and the closer $j$ is to $k$, the smaller is the error. Therefore, we use $\mathbf{t}_{\mathbf{v}}$ as an approximation for $\mathbf{h}$, up to  boundaries errors.

\section{Convex analysis}
\label{cv_analys}
In this appendix, we give the general convex analysis material relevant to our work. We consider a finite-dimensional Hilbert space $\mathcal{H}$ equipped with the inner product $\langle . , . \rangle$ and associated with the norm $\|.\|$.
       
\subsection{Proximity operator}
\paragraph{Definition:}
Let $\mathcal{F}$ be a real-valued function defined on $\mathcal{H}$.  $\mathcal{F}$ is proper if its domain, as defined by $\dom{\mathcal{F}} = \{\mathbf{x} \in \mathcal{H} / \mathcal{F}(x) < +\infty\}$, is non-empty. $\mathcal{F}$ is lower semicontinuous (LSC) if $\lim \inf_{\mathbf{x}\rightarrow \mathbf{x}_0}\mathcal{F}(\mathbf{x}) \geq \mathcal{F}(\mathbf{x}_0)$.
We define $\Gamma_0(\mathcal{H})$ as the set of proper LSC convex real-valued function defined on  $\mathcal{H}$.
For a function $\mathcal{F} \in \Gamma_0(\mathcal{H})$, the function $\mathbf{y} \rightarrow \frac{1}{2}\|\bm{\alpha}-\mathbf{y}\|^2+\mathcal{F}(\mathbf{y})$ achieves its minimum at a unique point denoted by $\prox_{\mathcal{F}}(\bm{\alpha})$, $(\forall \bm{\alpha} \in \mathcal{H})$ \cite{Moreau1965}; the operator $\prox_{\mathcal{F}}$ is the proximity operator of $\mathcal{F}$.
\paragraph{Examples:}
\begin{itemize}
\item let $\mathcal{C}$ be a convex closed set of $\mathcal{H}$. The indicator function of  $\mathcal{C}$ is defined as: 
\begin{equation}
	\bm{\iota}_{\mathcal{C}}(\mathbf{x}) = \left\lbrace \begin{matrix} 0, \; \text{if} \; \mathbf{x} \in \mathcal{C} \\
	+\infty, \; \text{otherwise}; 
\end{matrix}\right.\
\end{equation}	
it is clear from the definition that the proximity operator of $\bm{\iota}_{\mathcal{C}}$ is the orthogonal projector onto $\mathcal{C}$;
\item for $\mathcal{H} = \mathbb{R}$, $\lambda \in \mathbb{R}^+$ and $f: x \rightarrow \lambda |x|$, $\prox_f(y) = \ST_\lambda(y)$, where $\ST_\lambda$ denotes the soft-thresholding operator, with a threshold $\lambda$.
\end{itemize} 
\paragraph{Properties:}
\begin{itemize}
\item separability: if $\mathcal{H} = \mathcal{H}_1 \times \dots \times \mathcal{H}_n$, for $\mathcal{F} \in \Gamma_0(\mathcal{H})$ and if $\mathcal{F}(\mathbf{x}) = \mathcal{F}_1(\mathbf{x}[1])+\dots+\mathcal{F}_n(\mathbf{x}[n])$ where $\mathcal{F}_i \in \Gamma_0(\mathcal{H}_i)$, for $i=1\dots n$, then $\prox_\mathcal{F}(\mathbf{y}) = (\prox_{\mathcal{F}_1}(\mathbf{y}[1]),\dots,\prox_{\mathcal{F}_n}(\mathbf{y}[n]))$;
\item translation: for $\mathcal{F} \in \Gamma_0(\mathcal{H})$ and $\mathbf{a} \in \mathcal{H}$, we define $\mathcal{F}_{\mathbf{a}}(\mathbf{x}) = \mathcal{F}(\mathbf{x}-\mathbf{a})$; then $\prox_{\mathcal{F}_{\mathbf{a}}}(\mathbf{y}) = \mathbf{a}+\prox_\mathcal{F}(\mathbf{y}-\mathbf{a})$   
\end{itemize}

\subsection{Convex conjugate}
\paragraph{Definition:}
let $\mathcal{F}$ be a real-valued function defined on $\mathcal{H}$. The function $\mathcal{F}^*: \mathbf{y}\rightarrow \underset{\mathbf{x}}\max \langle \mathbf{x} , \mathbf{y} \rangle - \mathcal{F}(\mathbf{x})$ is the convex conjugate of $\mathcal{F}$; it is also known as the Legendre-Fenchel transformation of $\mathcal{F}$. 

\paragraph{Properties:} 
\begin{itemize}
\item Moreau identity: for $\mathcal{F} \in \Gamma_0(\mathcal{H})$ and $\lambda \in \mathbb{R}^*_+$, $\prox_{\lambda \mathcal{F}}(\mathbf{x})+\lambda\prox_{\frac{1}{\lambda}\mathcal{F}^*}(\frac{\mathbf{x}}{\lambda}) = \mathbf{x}$;
\item Fenchel - Moreau theorem: if $\mathcal{F} \in \Gamma_0(\mathcal{H})$, $\mathcal{F} = \mathcal{F}^{**}$.
\end{itemize}

\section{Minimization schemes}
\label{min_scheme}
This appendix details the practical resolution of the proposed method optimization problems. 

\subsection{Components estimation problem}
We consider the step 5 in the Alg. \ref{rca_algo}. 
If $\boldsymbol{\Phi}_s = \mathbf{I}_n$, the problem of estimating components takes the following generic form:
\begin{equation}
\underset{\mathbf{S}}\min \mathcal{F}(\mathbf{S})+\mathcal{G}_1(\mathcal{L}_1(\mathbf{S}))+\mathcal{H}(\mathbf{S}),
\end{equation}
with $\mathcal{F}(\mathbf{S}) = \sum_{i=1}^r \|\mathbf{w}_i\odot\mathbf{s}^{(c)}_i\|_1$, $\mathcal{G}_1 = \bm{\iota}_{\mathbb{R}_+^{n\times p}}$, $\mathcal{L}_1(\mathbf{S}) = \mathbf{S}\mathbf{A}$ and $\mathcal{H}(\mathbf{S}) = \frac{1}{2}\|\mathbf{Y}-\mathcal{M}(\mathbf{S})\|_F^2$ for some bounded linear operator $\mathcal{M}$.  

$\mathcal{F} \in \Gamma_0(\mathbb{R}^{n\times r})$, $\mathcal{G}_1 \in \Gamma_0(\mathbb{R}^{n\times p})$ and  $\mathcal{L}_1$ is a bounded linear operator. Moreover, $\mathcal{H}$ is convex, differentiable and has a continuous and Lipschitz gradient. This problem can be solved efficiently using the primal dual algorithms introduced in \cite{condat_vu} for instance. One only need to be able to compute $\lambda\mathcal{F}$ and $\alpha\mathcal{G}_1^*$ proximity operators, for some given positive reals $\lambda$ and $\alpha$ and $\mathcal{H}$'s gradient:
\begin{itemize}
\item $\prox_{\lambda\mathcal{F}}(\mathbf{S}) = (\hat{s}_{ij})_{1\leq i\leq n\atop 1\leq j\leq p}$, with $\hat{s}_{ij} = \ST_{\lambda\mathbf{w_j}[i]}(\mathbf{s_j}[i])$;
\item $\prox_{\alpha\mathcal{G}_1^*}(\mathbf{Z}) = \mathbf{Z} - (\mathbf{Z})_+$
\item $\nabla\mathcal{H}(\mathbf{S}) = - \mathcal{M}^*(\mathbf{Y}-\mathcal{M}(\mathbf{S}))$, where $\mathcal{M}^*$ is the adjoint operator of $\mathcal{M}$.
\end{itemize}
For an arbitrary dictionary $\boldsymbol{\Phi}_s$, we instead consider the following generic formulation of the problem:
 \begin{equation}
\underset{\mathbf{S}}\min \mathcal{G}_1(\mathcal{L}_1(\mathbf{S}))+\mathcal{G}_2(\mathcal{L}_2(\mathbf{S}))+\mathcal{H}(\mathbf{S}),
\end{equation}
where $\mathcal{G}_2(\mathbf{Z}) = \sum_{i=1}^r \|\mathbf{w}_i\odot\mathbf{Z}^{(c)}_i\|_1$ and $\mathcal{L}_2(\mathbf{S}) = [\boldsymbol{\Phi}_s\mathbf{s}^{(c)}_1,\dots,\boldsymbol{\Phi}_s\mathbf{s}^{(c)}_r]$.
One can use the algorithms suggested before and minimization will require the computation of  $\alpha\mathcal{G}_2^*$ proximity operator, for some given positive real $\alpha$ which is simply given by

$\prox_{\alpha\mathcal{G}_1^*}(\mathbf{Z}) = \mathbf{Z} - \widehat{\mathbf{Z}}$ ,with $\widehat{\mathbf{Z}}[i,j] = \ST_{\lambda\mathbf{w_j}[i]}(\mathbf{Z}[i,j])$.

\subsection{Coefficients estimation}
\label{app_coeff_est}
We consider the step 8 in the Alg. \ref{rca_algo}.
The problem takes the generic form:
\begin{equation}
\underset{\boldsymbol{\alpha}}\min \mathcal{J}(\boldsymbol{\alpha})\; \s.t.\; \|\boldsymbol{\alpha}[l,:]\|_0 \leq \eta_l, \; l=1\dots r, 
\label{coeff_est_gen}
\end{equation}
where $\mathcal{J}$ is convex, differentiable and has a continuous and Lipschitz gradient and $\boldsymbol{\alpha} \in \mathbb{R}^{r\times q}$. This problem is combinatorial and its feasible set is non-convex. For typical data sizes in image processing applications and tractable processing time, one can at best reach a "good" local optimum. There is an extensive literature on optimization problems involving the $\text{l}_0$ pseudo-norm. We propose an heuristic based on quite common ideas now and which appears to be convenient from a practical point of view. 
Let $\boldsymbol{\alpha}^*$ be a global minimum of Problem \ref{coeff_est_gen}. For a vector $\mathbf{M} \in \mathbb{R}^{r\times q}$, we define its support as 
\begin{equation}
\text{Supp}(\mathbf{M}) = \{(i,j)\in \llbracket 1,r \rrbracket \times \llbracket 1,q \rrbracket / |\mathbf{M}[i,j]| \geq 0 \}.
\label{supp}
\end{equation}
We note $\text{E}_{\boldsymbol{\alpha}^*}$ the set of $r\times q$ real matrices sharing the support of $\boldsymbol{\alpha}^*$:
\begin{equation}
\text{E}_{\boldsymbol{\alpha}^*} = \{\mathbf{M} \in \mathbb{R}^{r\times q} / \text{Supp}(\mathbf{M}) = \text{Supp}(\boldsymbol{\alpha}^*)\}.
\end{equation} 
$\text{E}_{\boldsymbol{\alpha}^*}$ is a vector space. In particular, $\text{E}_{\boldsymbol{\alpha}^*}$ is a convex set. 
Therefore, $\boldsymbol{\alpha}^*$ is a solution of the following problem: 
\begin{equation}
\underset{\boldsymbol{\alpha}}\min \mathcal{J}(\boldsymbol{\alpha})\; \s.t.\; \boldsymbol{\alpha} \in \text{E}_{\boldsymbol{\alpha}^*}.
\label{coeff_est_gen_supp}
\end{equation}
The proposed scheme is motivated by the idea of identifying approximately $\text{E}_{\boldsymbol{\alpha}^*}$ along with the iterative process. 
One can think of numerous algorithms to solve Problem \ref{coeff_est_gen_supp}, all involving the orthogonal projection onto $\text{E}_{\boldsymbol{\alpha}^*}$. We build upon the fast proximal splitting algorithm introduced in \cite{Beck2}. 
For a vector $\mathbf{u} \in \mathbb{R}^q$ we note $\boldsymbol{\sigma}$ a permutation of $\llbracket 1,q \rrbracket$ verifying $|\mathbf{u}[\boldsymbol{\sigma}(1)]| \geq \dots \geq |\mathbf{u}[\boldsymbol{\sigma}(q)]|$.
For an integer $k \leq q$, we define
\begin{equation}
\text{Supp}_k(\mathbf{u}) = \{i\in \llbracket 1,q \rrbracket/ |\mathbf{u}[i]| \geq |\mathbf{u}[\boldsymbol{\sigma}(k)]| \},
\end{equation}
Finally for a vector $\boldsymbol{\alpha} \in \mathbb{R}^{r\times q}$, we define the subspace
\begin{equation}
\text{E}_{k,\boldsymbol{\alpha}} =  \{\mathbf{M} \in \mathbb{R}^{r\times q} / \text{Supp}_k(\mathbf{M}[i,:]) = \text{Supp}_k(\boldsymbol{\alpha}[i,:]), i=1\dots r \}.
\end{equation}
The proposed scheme is given in Algorithm \ref{coeff_algo}. $f$ is a positive valued concave increasing function and  $\text{proj}_{\text{E}_{\floor{f(k)},\mathbf{U}_k}}(.)$ denotes the orthogonal projection onto $\text{E}_{\floor{f(k)},\mathbf{U}_k}$.

\begin{figure}
\begin{center}
\includegraphics[scale=0.50]{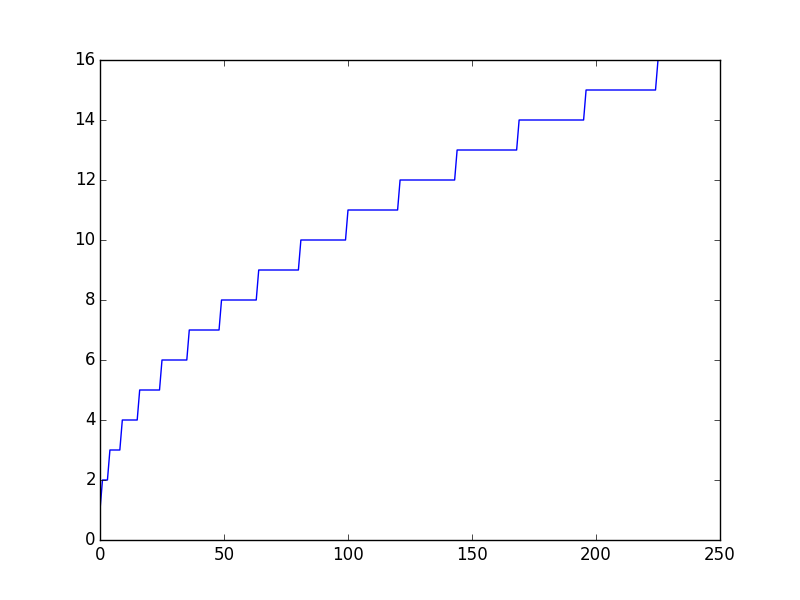}

\end{center}
\caption{Support size function; X axis: iteration index $k$ in Algorithm \ref{coeff_algo}; Y axis: $\floor{f(k)}$ for $f(x) = \sqrt{x}+1$}
\label{supp_size}
\end{figure}

\begin{algorithm*}[!htb]
\caption{Beck-Teboulle proximal gradient algorithm with variable proximity operator}
\begin{algorithmic}[1]
\label{coeff_algo}

\bigskip
\STATE \underline{Initialization}:  $\boldsymbol{\alpha}_0 =  0_{\mathbb{R}^{r\times q}}, \; \boldsymbol{\beta}_0 = \boldsymbol{\alpha}_0, \; t_0 = 1\; \text{res}_{-1} = 0, \; \text{res}_0 = 0, \; \text{tol}, \; k=0$
\STATE \underline{Minimization}
 \WHILE{$k<k_{\max}$ and $|(\text{res}_{k}-\text{res}_{k-1})/\text{res}_k|$}
 \STATE$\mathbf{U}_k = \boldsymbol{\beta}_k - \rho^{-1}\nabla\mathcal{J}(\boldsymbol{\beta}_k)$
 \STATE$\boldsymbol{\alpha}_{k+1} = \text{proj}_{\text{E}_{\floor{f(k)},\mathbf{U}_k}}(\mathbf{U}_k)$
 \STATE$t_{k+1} = \frac{1+\sqrt{4t_k^2+1}}{2}$
 \STATE$\lambda_k = 1 + \frac{t_k-1}{t_{k+1}}$
 \STATE$\boldsymbol{\beta}_{k+1} = \boldsymbol{\alpha}_{k} + \lambda_k(\boldsymbol{\alpha}_{k+1}-\boldsymbol{\alpha}_{k})$
 \STATE$\text{res}_{k+1} = \mathcal{J}(\boldsymbol{\beta}_k)$
 \STATE$k=k+1$
 \ENDWHILE
 
 \STATE {\bf Return:} $\boldsymbol{\alpha}_{k_{\text{stop}}}$.

\end{algorithmic}
\end{algorithm*}
The solution support size is constraint at step 5 and the size is gradually increased as shown in Fig. \ref{supp_size}. 

\begin{figure}
\begin{center}
\includegraphics[scale=0.25]{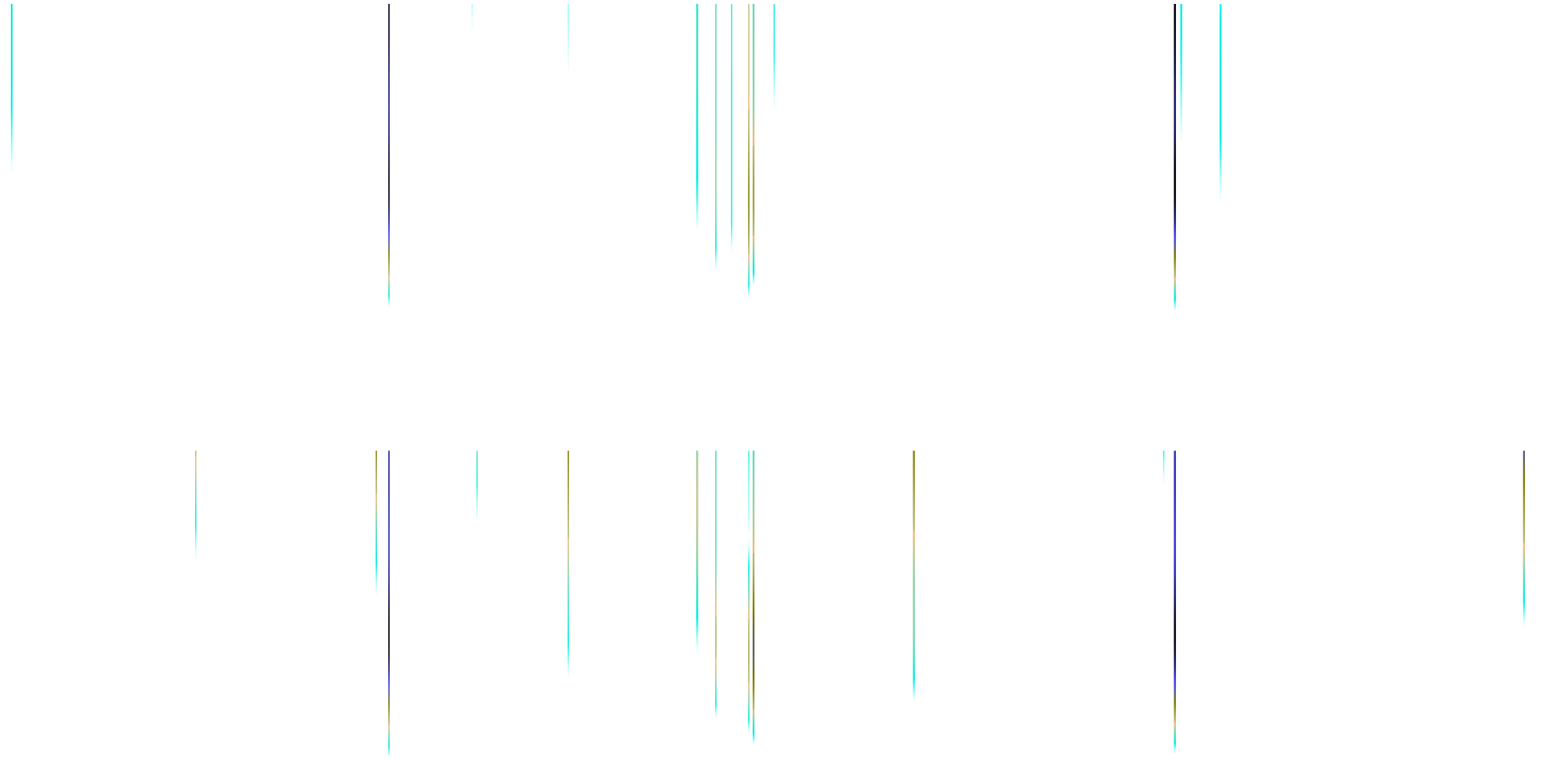}

\end{center}
\caption{Algorithm \ref{coeff_algo} main iterate evolution; X axis: $|\boldsymbol{\alpha}_{k+1}[0,:]|$ for the top image and $|\boldsymbol{\alpha}_{k+1}[1,:]|$ for the bottom image;\; Y axis: iterate index $k$}
\label{supp_trace}
\end{figure}

The convergence analysis this scheme is out of the scope of this paper. However, Fig. \ref{supp_trace} suggests that once an index is included in an iterate support, this index is included in all the forthcoming iterates supports. This implies that at each support size's step in Fig. \ref{supp_size}, the algorithm approximately solves a problem of the following form:     
\begin{equation}
\underset{\boldsymbol{\alpha}}\min \mathcal{J}(\boldsymbol{\alpha})\; \s.t.\; \boldsymbol{\alpha} \in \text{E}, 
\label{coeff_est_gen_loc}
\end{equation}
for a given subspace E, which is a convex problem.

This scheme can be viewed as an iterative hard thresholding \cite{blum}, with a decreasing threshold \cite{portilla}. Yet, it is quite easy to get an upper bound of the support size - related to the parameters $\eta_l$ in Problem \ref{coeff_est_gen} - from the data. Depending on the time one is willing to spend on the coefficients computation, this yields convenient choices for the function $f$.

\end{document}